\documentclass[final]{cvpr}

\usepackage{times}
\usepackage{epsfig}
\usepackage{graphicx}
\usepackage{amsmath}
\usepackage{amssymb}

\usepackage{makecell}
\usepackage{verbatim}
\usepackage{subfigure}
\usepackage{caption}
\usepackage{longtable}
\usepackage{setspace}
\usepackage{multirow}
\usepackage{booktabs}
\usepackage{bm}

\usepackage[pagebackref=true,breaklinks=true,colorlinks,bookmarks=false]{hyperref}

\def\cvprPaperID{5722} 
\def\confYear{CVPR 2021}

\begin{document}

\title{Styleverse: Towards Identity Stylization across Heterogeneous Domains}

\author{Jia Li, Jie Cao, JunXian Duan, Ran He\\
National Laboratory of Pattern Recognition, CASIA\\
Center for Excellence in Brain Science and Intelligence Technology, CAS\\
School of Artificial Intelligence, University of Chinese Academy of Sciences, Beijing, China\\}
\vspace{-18cm}
\twocolumn[{
\renewcommand\twocolumn[1][]{#1}%
\maketitle

\thispagestyle{empty} 
\begin{center}
    \centering
    \includegraphics[width=0.95\linewidth]{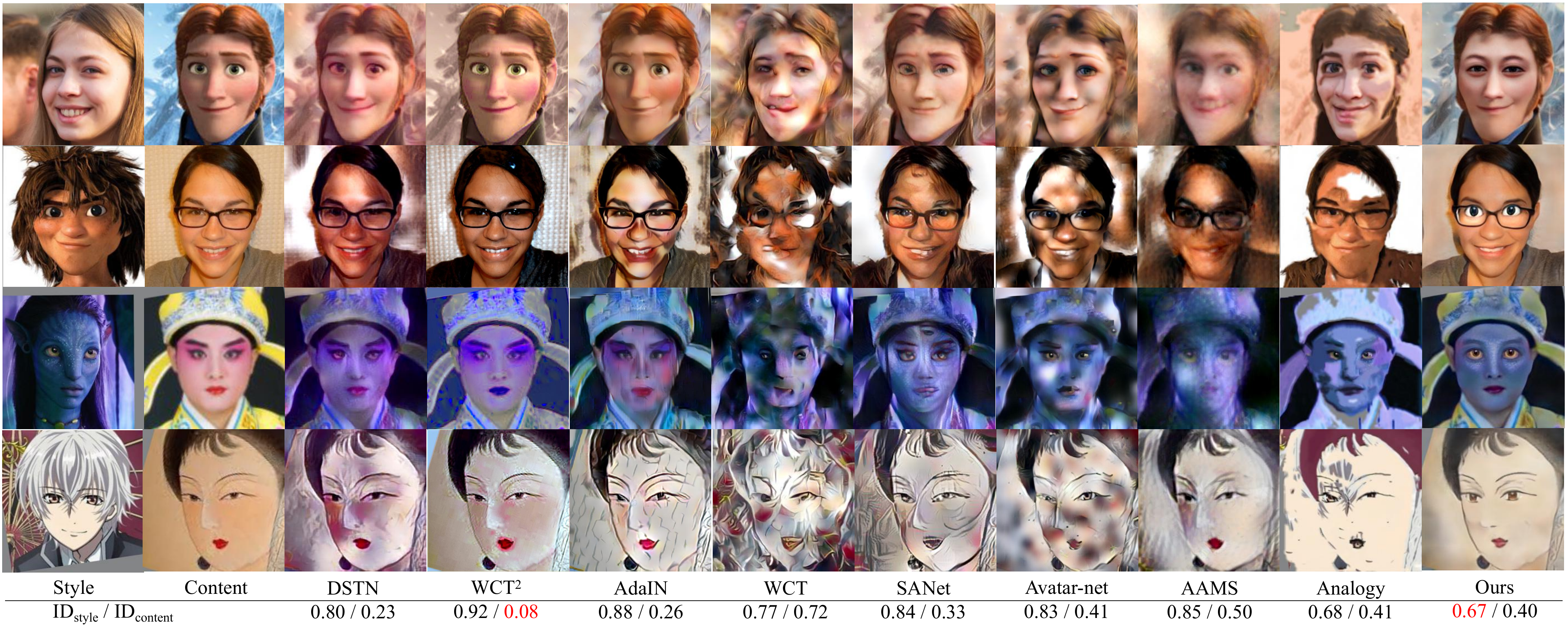}
    
    \vspace{-0.25cm}
    \captionof{figure}{Identity stylization results of Styleverse compared with state-of-the-art style transfer methods, \ie, DSTN \cite{dstn}, $WCT^{2}$ \cite{wct2}, AdaIN \cite{adain2017}, WCT \cite{WCT}, SANet \cite{sanet}, Avatar-net \cite{avatar}, AAMS \cite{aams} and deep analogy \cite{analogy}. Our Styleverse produces visually high-fidelity and topology-aware results based on the heterogeneous-domain styles. The identity distance \cite{arcface2019} between the styled identity and the original identity of the style image is indicated as $ID_{style}$,  and $ID_{content}$ represents the identity similarity of styled identity and content identity. Styleverse gets the best $ID_{style}$ score, while $WCT^{2}$ gets the best $ID_{content}$ score but with the worst $ID_{style}$ score. Note that Styleverse focuses on identity stylization rather than identity swapping, demonstrated by the last score pair, \ie, the stylized identity is still more similar with that of the source content rather than the referenced face.}
    \label{fig:begin}
\end{center}
}]
\begin{abstract}
   We propose a new challenging task namely IDentity Stylization (IDS) across heterogeneous domains. IDS focuses on stylizing the content identity, rather than completely swapping it using the reference identity. We use an effective heterogeneous-network-based framework $Styleverse$ that uses a single domain-aware generator to exploit the Metaverse of diverse heterogeneous faces, based on the proposed dataset FS13 with limited data. FS13 means 13 kinds of Face Styles considering diverse lighting conditions, art representations and life dimensions. Previous similar task, \eg, image style transfer can handle textural style transfer based on a reference image. This task usually ignores the high structure-aware facial area and high-fidelity preservation of the content.  However, Styleverse intends to controllably create topology-aware faces in the Parallel Style Universe, where the source facial identity is adaptively styled via AdaIN guided by the domain-aware and reference-aware style embeddings from heterogeneous pretrained models. We first establish the IDS quantitative benchmark as well as the qualitative Styleverse matrix. Extensive experiments demonstrate that Styleverse achieves higher-fidelity identity stylization compared with other state-of-the-art methods. 
\end{abstract}

\section{Introduction}
\label{sec:intro}
\begin{figure*}[t]
\begin{center}
\includegraphics[width=0.8\linewidth]{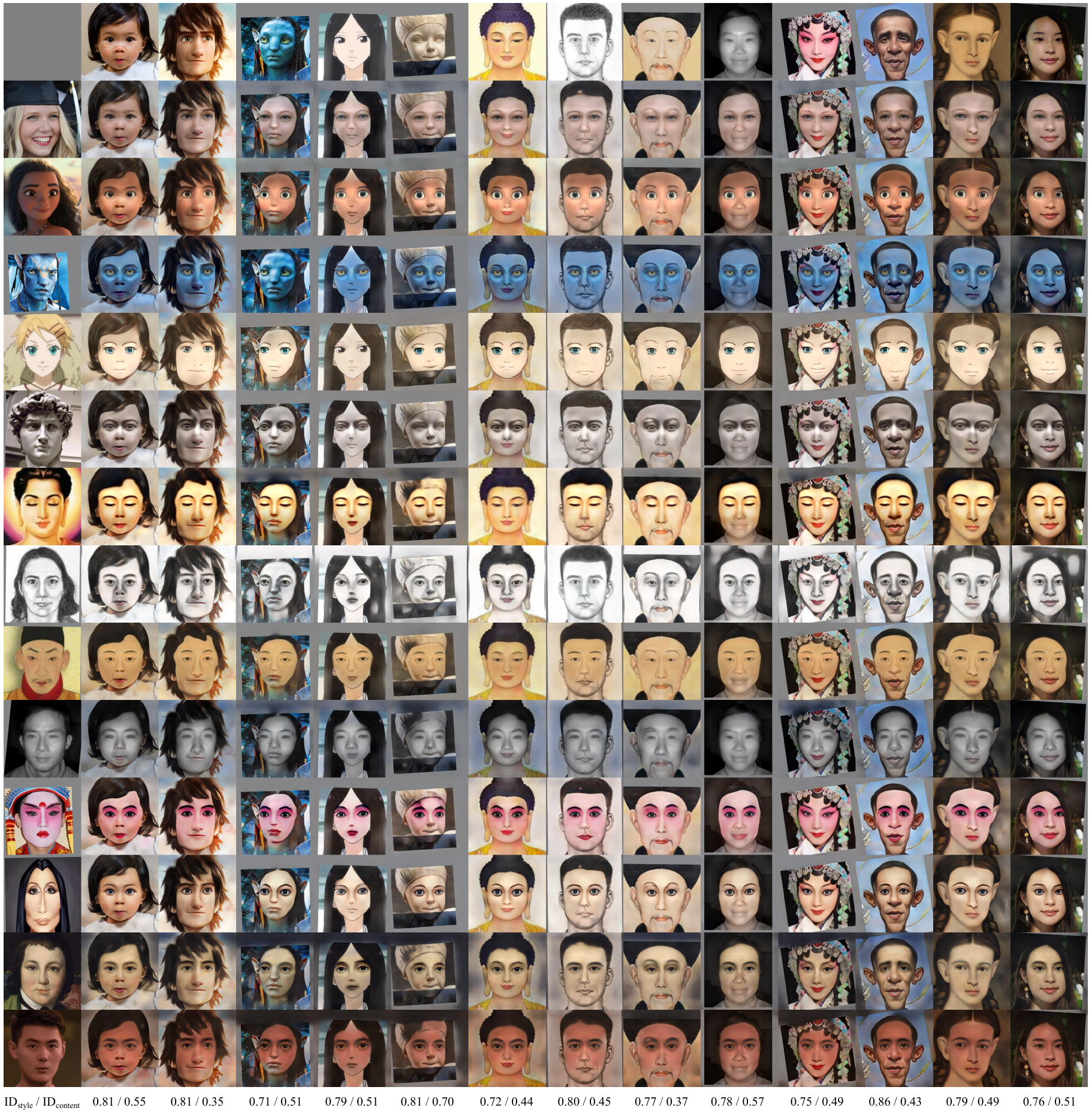}
\end{center}
\vspace{-18pt}
   \caption{We build a heterogeneous domain-aware identity stylization model to generate characters in the Parallel Style Universe (PSU) using a single generator. The $i=0\ th$ row and $j=0\ th$ column represent the content and style images, and all these generated faces constitute the Styleverse matrix. We calculate the identity distance \cite{arcface2019} for each result with the corresponding style and content image respectively, \eg, $ID_{style}=\dfrac{1}{N}\sum_{i=1}^{N=13}ID_{i,1}^{i,0}=0.81$ and $ID_{content} = \dfrac{1}{N}\sum_{i=1}^{N=13}ID_{i,1}^{0,1}=0.55$ in col 1. Overally, $ID_{style}=0.78$ and $ID_{content}=0.49$, which indicates the main difference between identity stylization and identity swapping.}
\label{fig:matrix}
\vspace{-10pt}
\end{figure*}
\begin{figure}[h]
\begin{center}
\includegraphics[width=1\linewidth]{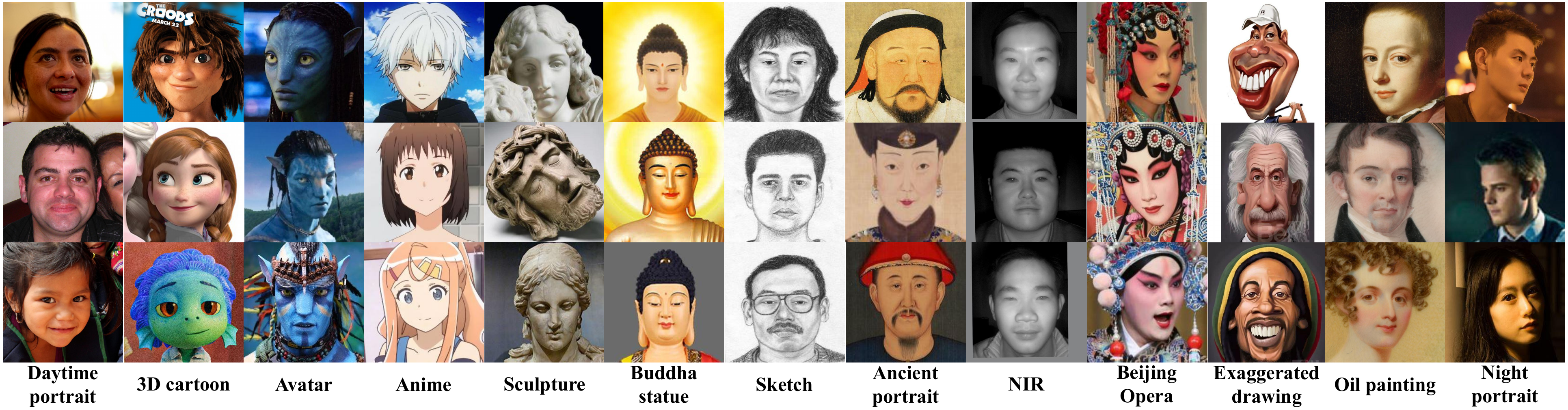}
\end{center}
\vspace{-13pt}
   \caption{We establish a new heterogeneous dataset $FS13$ for universal face style transfer, including portraits with different lighting conditions (daytime, night, NIR), art representations (2D or 3D colored cartoon, sketch, sculpture, Beijing Opera, exaggerated or oil painting) and life dimensions (Buddha statue, Avatar,  modern or ancient people).}
\label{fig:da13}
\vspace{-10pt}
\end{figure}
Style transfer is assumed that there exists a single generator that can generate diverse styles for certain content \cite{adain2017,WCT,wct2,dstn}. These methods mainly transfer the textural style and the styled faces have obvious artifacts that destroy the facial topology, as shown in Figure \ref{fig:begin}. As for the proposed identity stylization (IDS) task, it is supposed to consider the highly topology-aware facial area and high-fidelity heterogeneous-domain style transfer, as shown in the last column of Figure \ref{fig:begin}. Furthermore, if there are $N$ Parallel Style Universes, and each PSU has $M$ individuals, it will synthesize about $(MN)^{2}$ new-created results. We call this single generator as $Styleverse$. Another popular concept is Metaverse \cite{snow}, which connects the real world and virtual world by means of VR and AR. Similarly, Styleverse connects diverse PSUs as well. However, Styleverse is a challenging and lightweight Artificial Intelligence task, because the PSU styles are infinite, but the universal generator is only one. In this paper, we focus on the human face with significant identity, and exploit Styleverse in the image space, as shown in Figure \ref{fig:matrix}.

Our Styleverse studies heterogeneous image generation (HIG) that has made a great progress in recent years. Previous HIG methods are proposed to deal with two-domain dual generation. For example, DVG \cite{fu2019dual} proposes a dual variational generator to synthesize jointly paired heterogeneous images. UGATIT \cite{ugatit} studies VIS-anime and VIS-oil painting, but trains models respectively for these two tasks. Additionally, if achieving $N$-domain IDS, UGATIT needs to train  $C_{N}^{2}$ times, which will rely on a large amount of computation. The same goes for \cite{cyclegan,discogan, dualgan}. We propose Styleverse to create the 2D styled faces cross different heterogeneous domains, achieving an universal generator to represent diverse parallel style universes.
Our contributions are three folds as follows. 
\begin{itemize}
\item We first study the identity stylization, which is an universal face style transfer task. Inspired by Metaverse, we propose a new concept $Styleverse$ which uses an universal framework to deal with the challenging and meaningful problem. Furthermore, we establish a new dataset FS13 with several heterogeneous domains.  
\item We propose an effective heterogeneous-network-based IDS framework. By considering high-fidelity style and content, our approach achieves high-quality and topology-aware identity stylization in diverse parallel style universes using only one generator. Moreover, Styleverse can control the textural details based on the referenced PSU images.
\item We analyse different Styleverse variants, and compare with state-of-the-art style transfer methods. We establish the first IDS quantitative benchmark and qualitative Styleverse matrix.
\end{itemize}
\begin{figure}[t]
\begin{center}
\includegraphics[width=1\linewidth]{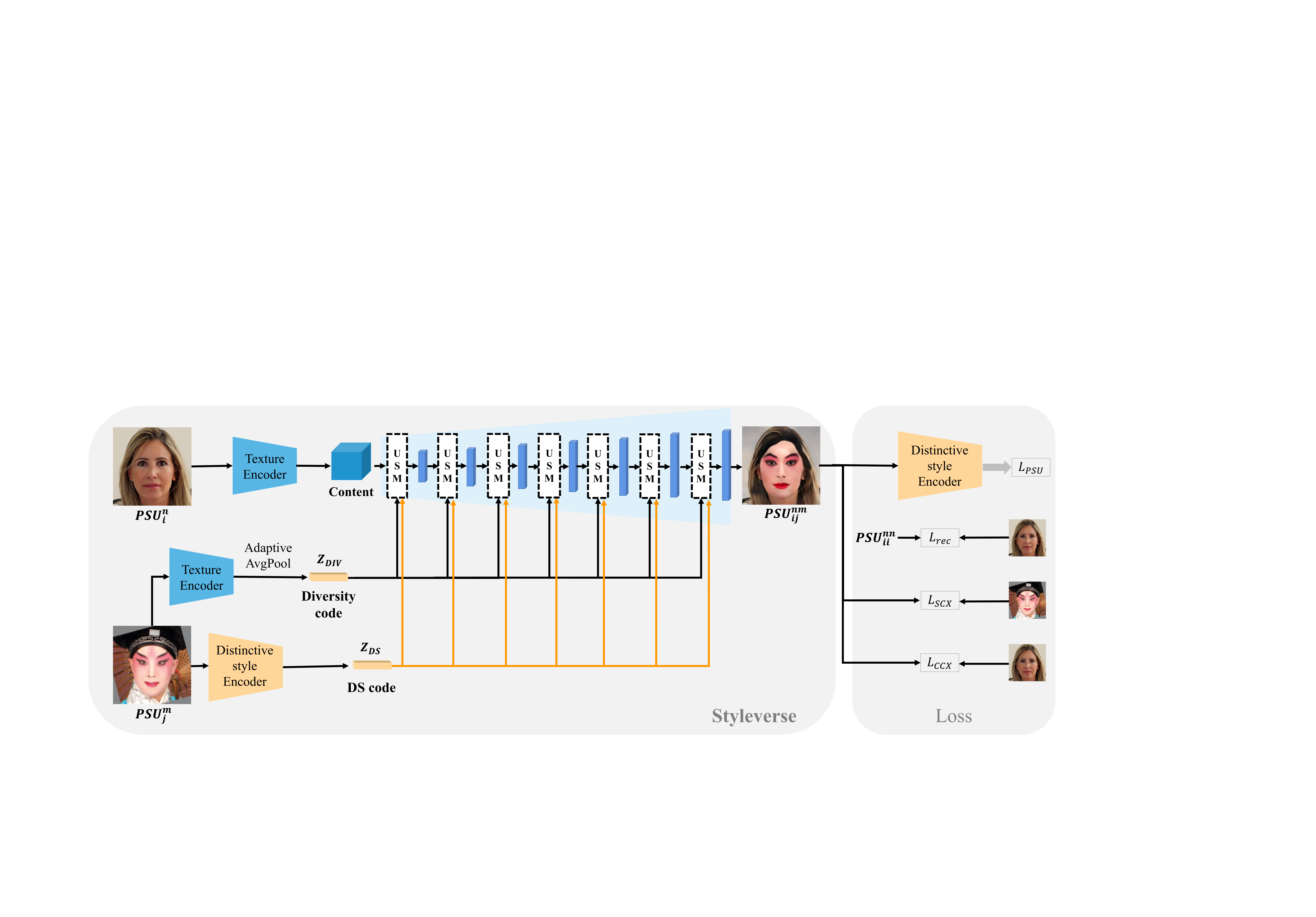}
\end{center}
\vspace{-13pt}
   \caption{Simple and effective framework of Styleverse for identity stylization across heterogeneous domains. We propose to apply heterogeneous network architectures for Styleverse. The content feature is extracted by the texture encoder VGG. The stlyle codes consist of distinctive style code and diversity code, which control the global PSU style and instance reference style. Texture encoder and distinctive style encoder mean the pretrained VGG19 \cite{vgg2015} and LightCNN \cite{lightcnn}, whose architectures are heterogeneous. The universal style modulation (USM) module is used to integrate the DS code and diversity code, and control the content feature transformation.}
\label{fig:psu}
\end{figure}
\section{Related Work}
\subsection{Style-code-based methods}
StarGAN v2 \cite{stargan2} consists of four modules including generator, mapping network, style encoder and discriminator. This method uses AdaIN \cite{adain2017} to control different face styles, e.g., gender, identity, pose and expression in the real-world domain. StyleGANs \cite{stylegan19, stylegan2_2020, stylegan3} synthesize high-fidelity images with noise modulation. UGATIT \cite{ugatit} modulates the generator using the learned domain-aware style code. NICE-GAN \cite{nicegan} reuses the discriminator for encoding to implement unsupervised image-to-image translation based on AdaILN \cite{ugatit}. Some face-swapping methods \cite{faceshifter_2020, faceinpainter, simswap} conduct identity transfer based on the identity embedding of Arcface \cite{arcface2019}. \cite{fudis} proposes an unsupervised face alignment method with AdaIN to disentangle facial shapes and identities. Styleverse also controls face styles with the guidance of domain-aware heterogeneous code and individual-aware textural code.

\subsection{Heterogeneous image generation}
As for NIR-VIS HIG, Fu et al. \cite{fu2019dual, fu2021dvg} achieve dual heterogeneous face representation based on variational autoencoder (VAE) \cite{vae}, which promotes heterogeneous face recognition. As for Sketch-VIS HIG, \cite{skph} utilizes a dual-transfer sketch-photo synthesis framework considering both inter-domain and intra-domain transfer process. As for Grayscale-VIS HIG, \cite{memopaint} proposes an adaptive painting model based on color memory. \cite{ugatit} achieves unsupervised translation of oil-VIS and anime-VIS HIG with Adaptive Layer-Instance Normalization. Different from above methods that train customized GAN or VAE for two specific heterogeneous domains, our IDS task designs an universal generator to carry out the multi-domain HIG.

\subsection{Heterogeneous datasets}

NIR-VIS datasets include CASIA NIR-VIS 2.0 \cite{casia_nir_vis2}, Oulu-CASIA NIR-VIS \cite{ouluCASIA} and the BUAA-VisNir Face \cite{BUAA_VisNir} databases. Sketch-VIS datasets include CUHK Face Sketch FERET (CUFSF) \cite{sketch_vis}. More sketch datasets focus on sketch-photo generation and sketch face recognition, e.g., IIIT-D Viewed Sketch database \cite{IIITD},  XM2VTS database \cite{XM2VTS}. \cite{ugatit} proposes selfie2anime dataset with diverse poses. IIIT-CFW \cite{cfw} contains some exaggerated drawings of famous people. Different from above datasets, we establish a novel heterogeneous dataset including more diverse styles to study the mutual IDS among a variety of styles.



\section{Problem setting}
\subsection{IDS}
Given a target PSU reference image $PSU_{j}^{m}$ and a source PSU content $PSU_{i}^{n}$ where $\{i,j\} \in \{1,2,...,N\}$, $\{m,n\} \in \{1,2,...,M\}$, we learn an universal knowledge leveraging a single generator $\mathbf{G}$ to achieve identity stylization between arbitrary two domains in N domains. This is fomulated as:
\begin{equation}
PSU_{ij}^{nm}=\mathbf{G}(PSU_{i}^{n}, PSU_{j}^{m})
\end{equation}

There are four significant features of the defined IDS problem.
\begin{itemize}
\item The IDS model is an universal framework for diverse heterogeneous-domain image translation in an unsupervised manner. Moreover, it considers the diversity in the specific domain.  
\item The IDS model is lightweight. There is only one generator in Styleverse. Other priors or loss constraints are obtained from the pretrained models.
\item The training data is limited. The proposed dataset FS13 has 100 samples for each PSU domain. IDS is a few-shot learning problem.
\item IDS considers highly topology-aware facial area and high-fidelity style transfer. Moreover, the stylized identity is still more similar with that of the source content rather than the referenced face.
\end{itemize}

\subsection{FS13 dataset}
To study the challenging IDS problem, we propose FS13 dataset, where there are 13 PSU heterogeneous domains considering abundant lighting conditions, art representations and life dimensions, as shown in Figure \ref{fig:da13}.
\begin{itemize}
\item \textbf{Daytime portrait}. We select face images with variable poses, expressions and genders from FFHQ \cite{stylegan19}. 

\textbf{Night portrait}. We crawl male and female night portraits with diverse lighting conditions at night.

\textbf{NIR}. We randomly select some identities from CASIA NIR-VIS 2.0 \cite{casia_nir_vis2}.
\item \textbf{3D cartoon}. Some character faces come from LUCA, Frozen, Tangled, and The Croods.

\textbf{Anime}.  We select some girl faces in selfie2anime dataset \cite{ugatit}, and crawl some boy faces.

\textbf{Sculpture}. There are some famous sculptures of Angel, Madonna, Knight and David. 

\textbf{Sketch}.  We randomly select some sketch faces from CUHK Face Sketch FERET (CUFSF) \cite{sketch_vis} and XM2VTS database \cite{XM2VTS}.

\textbf{Beijing Opera}. We collect some makeup faces of female and male roles in Beijing Opera. 

\textbf{Exaggerated drawing}. We randomly select some exaggerated drawings from IIIT-CFW \cite{cfw} and enhance them to higher resolution using \cite{gfpgan}.

\textbf{Oil painting}. We select male and female oil paintings from MetFaces \cite{MetFaces}.
\item \textbf{Buddha statue}. We collect different Buddha statues (only for research) with various skin colors and backgrounds.

\textbf{Avatar}. We collect the faces of characters with different poses, expressions and lighting conditions in $Avatar$.

\textbf{Ancient portrait}. We collect some ancient drawings of emperors, officials, poets and scholars.
\end{itemize}
\begin{figure}[t]
\begin{center}
\includegraphics[width=1\linewidth]{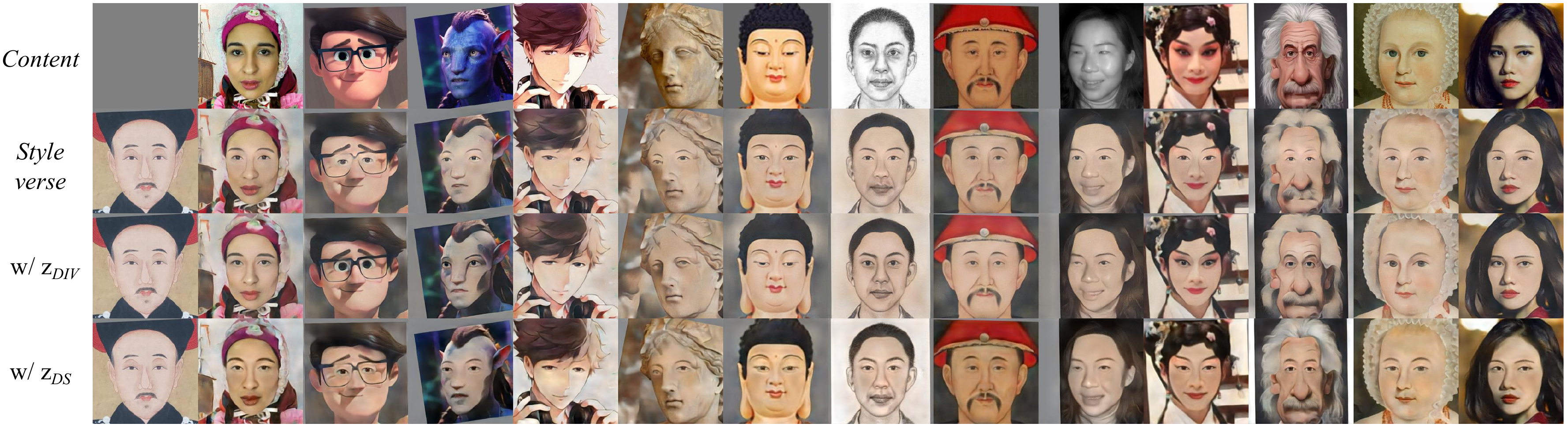}
\end{center}
\vspace{-13pt}
   \caption{Visual comparison of Styleverse variants, where only w/ $z_{DIV}$ mainly captures the texture style from the target PSU image, but the significant facial components, e.g., eyes, are still maintained as the source domain. While only w/ $z_{DS}$ preserves more source texture information of the content (row 1), \eg, Buddha statue. Styleverse achieves better texture-aware IDS compared with others.}
\label{fig:DS}
\end{figure}
\begin{figure}[ht]
\begin{center}
\includegraphics[width=1\linewidth]{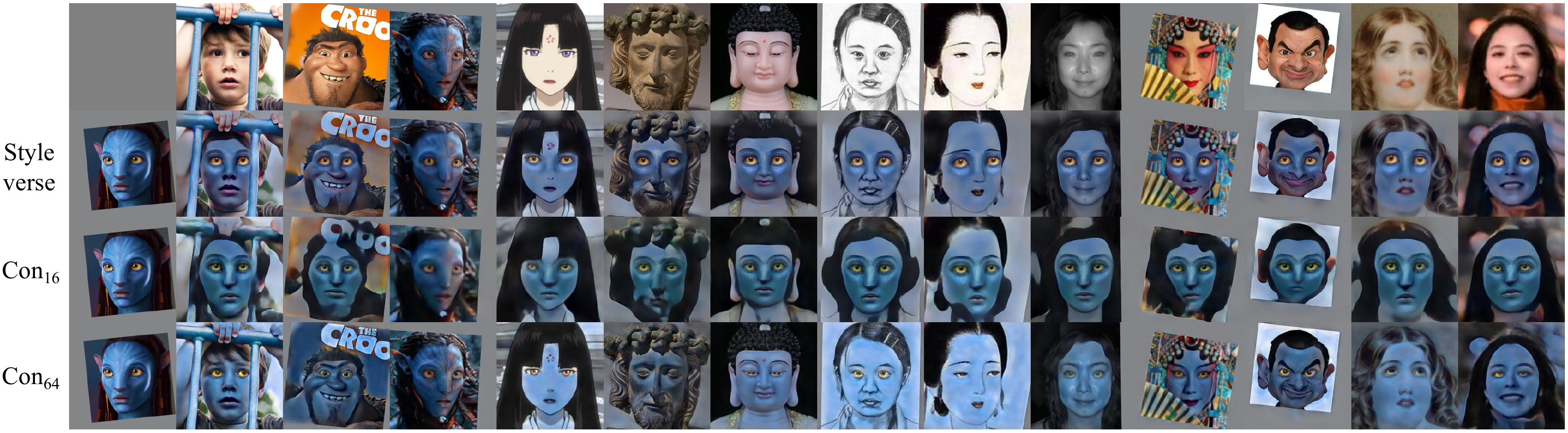}
\end{center}
\vspace{-13pt}
   \caption{Visual comparison of Styleverse variants with the same CCX weight. $Con_{16}$ means that the content feature size is $16\times16$. It's hard for $Con_{16}$ to preserve the source content. While $Con_{64}$ preserves excessive source style, \eg, facial component shape. Styleverse achieves more vivid IDS results.}
\label{fig:1664}
\vspace{-10pt}
\end{figure}
\section{Styleverse}
\subsection{Approach}
 There are some good-performance GAN-based image synthesis methods, e.g., StyleGAN \cite{stylegan19}, StarGAN \cite{stargan2}, UGATIT \cite{ugatit}, where discriminators are used to promote the domain-specific generator close to the target distribution. However, it is hard for one generator to deal with diverse heterogeneous domains, especially considering the high topology-sensitive face. If training specific discriminator for each domain, it will be computation-consuming. Lightweight and universal intelligence is the future direction of research. Different from the above GAN-based approaches, we incorporate heterogeneous domain-aware prior, texture prior from the pretrained models with the universal generator $\mathbf{G}$. Moreover, we polish Styleverse with global and local style preservation losses,  described as follows. 

As shown in Figure \ref{fig:psu}, given a source image $PSU_{i}^{n} \in \mathbb{R}^{3\times H \times W}$, we extract the content prior $Con_{i}^{n} \in \mathbb{R}^{C\times CH \times CW}$ using pretrained VGG \cite{vgg2015}. At the same time, we extract the diversity code $z_{DIV}$ from a target reference image $PSU_{j}^{m} \in \mathbb{R}^{3\times H \times W}$ by VGG extractor with adaptive average pooling. 
\begin{figure}[t]
\begin{center}
\includegraphics[width=1\linewidth]{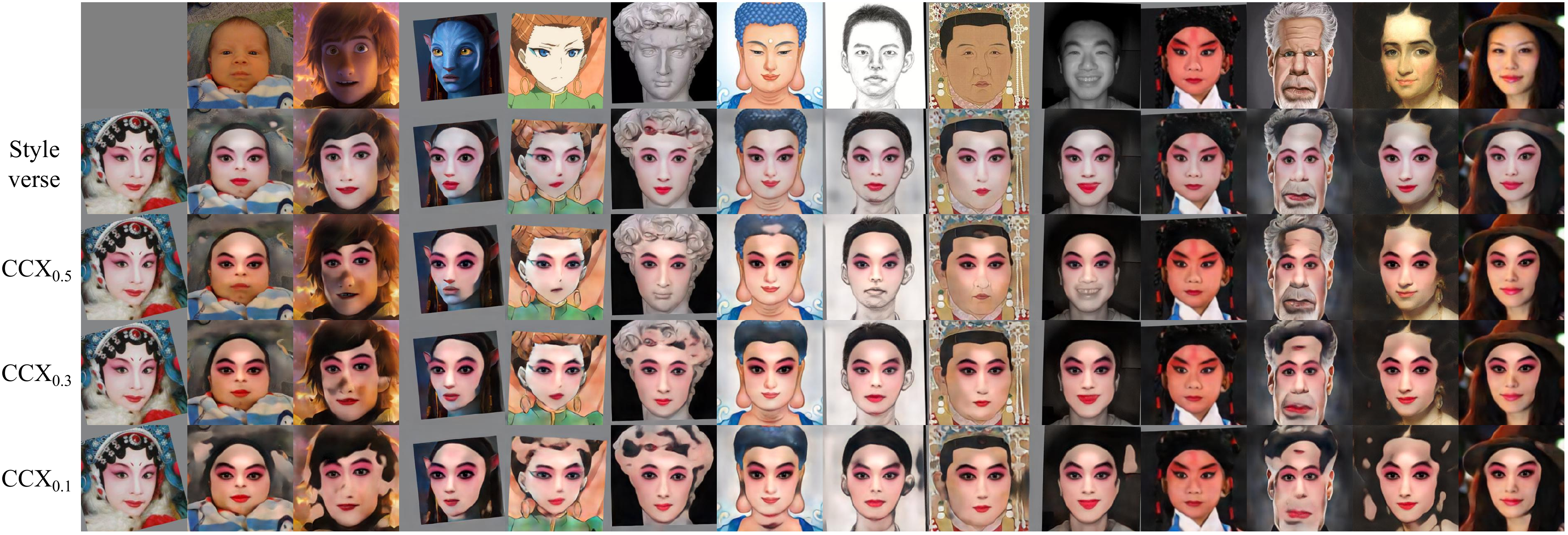}
\end{center}
\vspace{-13pt}
   \caption{Visual comparison of Styleverse variants with the same content feature size, \ie, $Con_{32}$. CCX means the contextual loss of the content. Lower CCX weight allows more obvious IDS close to the style identity, but more distorted source content. Styleverse uses $CCX_{0.5}$ and has a stable IDS performance. Note that the last 3 rows do not integrate $z_{DIV}$.}
\label{fig:weight}
\vspace{-10pt}
\end{figure}
\begin{figure}[ht]
\begin{center}
\includegraphics[width=1\linewidth]{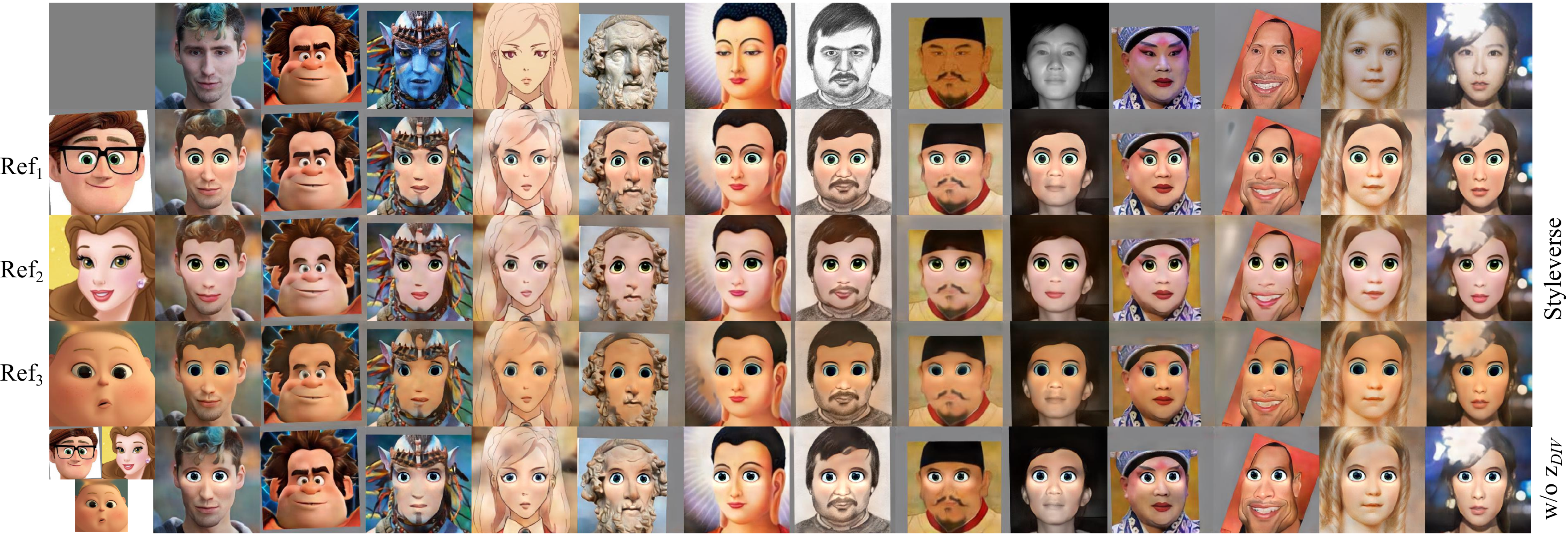}
\end{center}
\vspace{-13pt}
   \caption{Diverse IDS results of Styleverse. While w/o $z_{DIV}$ can not capture the significant style feature of different target references of 3D cartoon domain. Styleverse controls the skin colors and component features based on diverse reference faces.}
\label{fig:div}
\vspace{-10pt}
\end{figure}

As for the distinctive style encoder, we use the pretrained LightCNN \cite{lightcnn} to extract the significant heterogeneous domain-aware prior. LightCNN introduces the max-feature-map (MFM) to conduct competitive feature activation, where the architecture is heterogeneous compared with Conv-BN-ReLU based VGG. The heterogeneous style classification based on Binary Cross Entropy loss is fomulated as:
\begin{equation}
\mathcal{L}_{cls}=-\frac{1}{NM}\sum_{t=1}^{NM}y_{PSU}^t\log (\hat{y}_{PSU}^t),
\end{equation}
where $\hat{y}_{PSU}^t$ is the output of LightCNN for $t$ th sample, and $y_{PSU}^t$ is the PSU label associated with it.

In the hierarchical universal style modulation (USM) module, we map $z_{DIV}$ and $z_{DS}$ to the integrated latent vector $z_{style}$ by means of a fully connected layer for each feature level of $\mathbf{G}$. Then we inject $z_{style}$ to StyledConv block of StyleGAN \cite{stylegan2_2020} by implementing multi-scale AdaIN. It is formulated as:
\begin{equation}
AdaIN(Con^{i},\gamma ^{i} _{style},\beta^{i} _{style})=\gamma _{style}^{i}\odot  \dfrac{Con^{i}-\mu^{i}}{\sigma ^{i}} +\beta _{style}^{i},
\end{equation}
where  $Con^{i} \in \mathbb{R}^{C_{Con}^{i}\times H^{i} \times W^{i}}$ is the spatial embedding of the modulated information in $USM_{i}$. $\mu^{i}$ and $\sigma^{i}$ are the means and standard deviations of $Con^{i}$. $\gamma _{style}^{i}$ and $\beta _{style}^{i}\in \mathbb{R}^{C_{Con}^{i}\times H^{i}\times W^{i}}$ based on $z_{style}$ are used to conduct the instance denormalization.

\subsection{Objective Function}
Inspired by semi-supervised learning strategy \cite{ugatit}, we use the reconstruction loss to penalize the pixel-level distances between $PSU_{ii}^{nn}$ and $PSU_{i}^{n}$ in the case of the same PSU. It is formulated as
\begin{equation}
\mathcal{L}_{rec}=\begin{cases}\dfrac{1}{2}\left\|PSU_{ij}^{nm}-PSU_{i}^{n}\right\| _{2}^{2} \quad if \ i=j, n=m\\ \quad \quad \quad 0   \quad \quad \quad \quad \text otherwise\end{cases}.
\end{equation}

We use heterogeneous domain-aware prior consistency loss on $PSU_{ij}^{nm}$ for IDS face synthesis as follows. Specifically, we calculate the distance between the LightCNN heterogeneous-domain embeddings of $PSU_{ij}^{nm}$ and $PSU_{j}^{m}$ via
\begin{equation}
\mathcal{L}_{PSU}=1 -\langle z_{DS}(PSU_{ij}^{nm}) ,z_{DS}\left(PSU_{j}^{m}\right) \rangle,
\end{equation}
where $\langle\cdot, \cdot\rangle$ means cosine similarity.

\begin{figure}[ht]
\begin{center}
\includegraphics[width=1\linewidth]{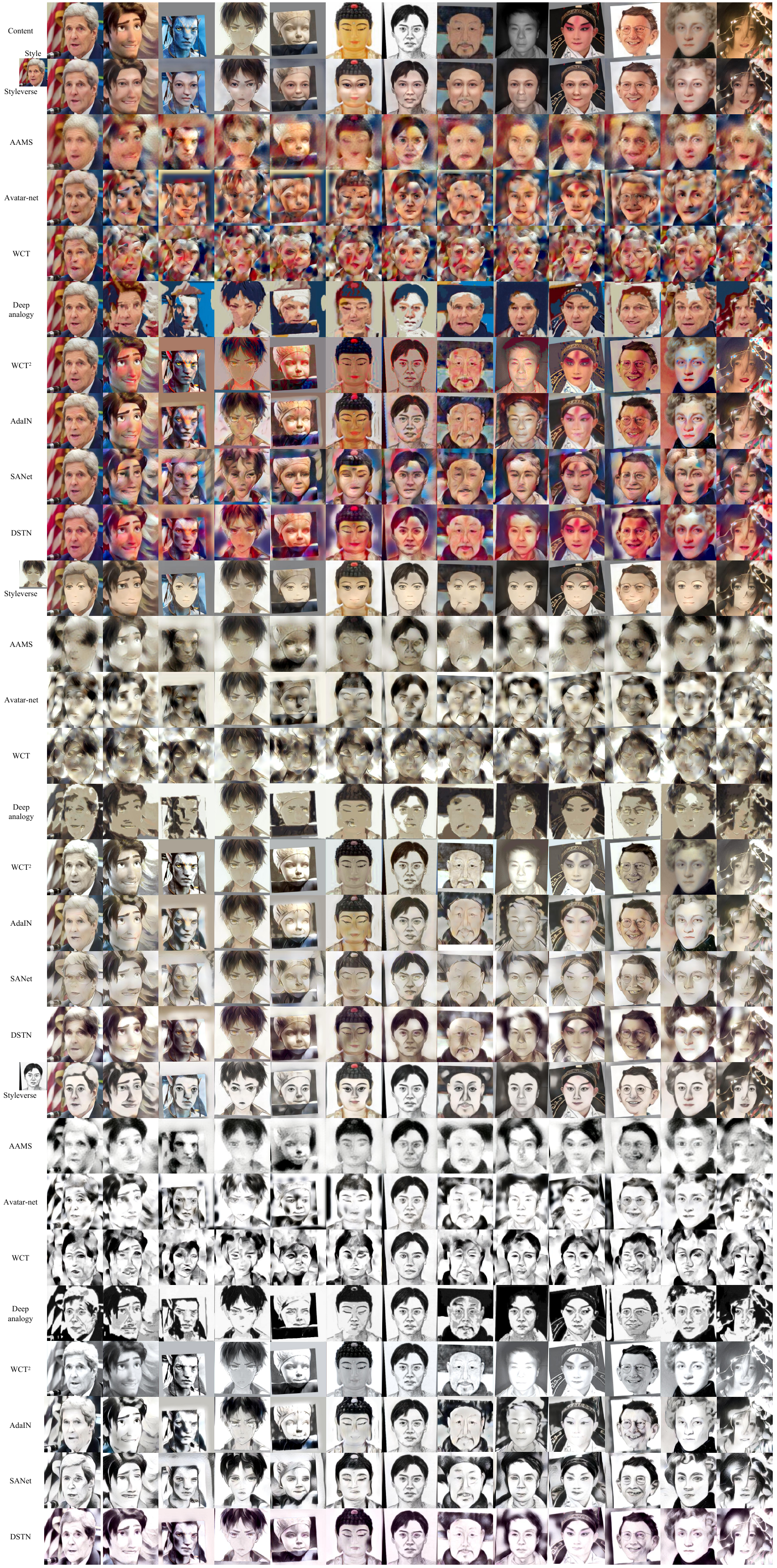}
\end{center}
\vspace{-13pt}
   \caption{Visual IDS comparison of Styleverse, \cite{aams}, \cite{avatar}, \cite{WCT}, \cite{analogy}, \cite{wct2}, \cite{adain2017}, \cite{sanet} and \cite{dstn}. We take daytime portrait, anime and sketch face as the examples. The results of AAMS \cite{aams}, Avatar-net \cite{avatar} and WCT \cite{WCT} are ambiguous. The results of deep analogy \cite{analogy} and SANet \cite{sanet} are not topology-aware enough, \ie, the facial topology is distorted. While the IDS ability of $WCT^{2}$ \cite{wct2}, AdaIN \cite{adain2017} and DSTN \cite{dstn} is limited, \ie, the shape and texture of facial components are preserved excessively. Styleverse achieves high-fidelity IDS results.}
\label{fig:very_big}
\end{figure}

$\mathcal{L}_{PSU}$ is competent to transferring critical domain style as a global style constraint. To adaptively reconstruct the IDS result considering the contextual similarity, we utilize contextual loss \cite{cx2018} to measure the feature similarity between $PSU_{ij}^{nm}$ and $PSU_{j}^{m}$ for contextual-aware style transfer. This loss helps to produce comparable results without using discriminator-based GANs \cite{cx2018}. It is formulated as
\begin{equation}
\mathcal{L}_{SCX}=-\log ( CX(F_{vgg}^{l} (PSU_{ij}^{nm}),F_{vgg}^{l}(PSU_{j}^{m}))),
\end{equation}
where $l$ means  \emph{relu}\{3$\_$2, $4\_2$\} layers of the pretrained VGG19 network. Similarly, we add an content contextual loss between $PSU_{ij}^{nm}$ and $PSU_{i}^{n}$ for content preservation.
\begin{equation}
\mathcal{L}_{CCX}=-\log ( CX(F_{vgg}^{l} (PSU_{ij}^{nm}),F_{vgg}^{l} (PSU_{i}^{n}))).
\end{equation}

The total loss of Styleverse is as follows
\begin{equation}
\begin{split}
    \mathcal{L}_{\emph{Styleverse}}= \lambda _{PSU}\mathcal{L}_{PSU} + \lambda _{rec}\mathcal{L}_{rec}  + \\\lambda _{CCX}\mathcal{L}_{CCX} + \lambda _{SCX}\mathcal{L}_{SCX}.
\end{split}
\label{con:8}
\end{equation}

\section{Experiment}
\subsection{Implementation Details}
Styleverse is trained with the proposed FS13 dataset where $N=13, M=100$. We align and crop the source and target faces using five point landmarks \cite{gfpgan}. All samples are resized to 256$\times$256 resolution. We set $Con_{i}^{n} \in \mathbb{R}^{512\times 32 \times 32}$, $z_{DS}$, $z_{DIV}$ and $z_{style} \in \mathbb{R}^{512\times 1}$. There  are 8 USM modules. In Equation \ref{con:8}, $\lambda_{PSU}=\lambda_{SCX}=1$, $\lambda_{rec}=100, \lambda_{CCX}=0.5$. We randomly choose two domains from FS13 as the content and style input with batchsize 4.


\subsection{Analysis of Styleverse}
\paragraph{PSU texture transfer} 
As shown in Figure \ref{fig:DS}, the texture style of w/ $z_{DS}$ is not consistent with the reference style image, \eg, the daytime and ancient portraits, Buddha statue, and oil painting. And only w/ $z_{DIV}$ has poor performance on transferring distinctive facial manner of the style image. Styleverse considers both domain-level and instance-level IDS by integrating $z_{DS}$ and $z_{DIV}$.
\paragraph{PSU content dimension} 
As shown in Figure \ref{fig:1664}, the low-level details of $Con_{16}$ are limited, which results in the distorted background. Furthermore, if preserving a larger-size input content feature, the created Avatars are not vivid enough, \eg, eyes and delicate skin of the alien life. Styleverse takes both high-fidelity low-level scene and high-level style into account.

\begin{table*}[h]
\caption{Quantitative evaluation on FS13 with FID, KID$\times100$, NIQE, $ID_{style}$ and $ID_{content}$. The red values mean the top-1 scores for each row in the specific metric, among Styleverse, AAMS \cite{aams}, Avatar-net \cite{avatar}, WCT \cite{WCT}, $WCT^{2}$ \cite{wct2}, AdaIN \cite{adain2017}, SANet \cite{sanet} and DSTN \cite{dstn}. Note that the first column means the style domains. Our FID scores are higher than that of SANet \cite{sanet}, because IDS of Styleverse maintains the content using $\mathcal{L}_{CCX}$. Note that deep analogy \cite{analogy} is time-consuming, so we only compare it in Figure \ref{fig:very_big}.}
\vspace{-10.pt}
\scalebox{0.295}{
\begin{tabular}{|c|cccccccc|cccccccc|cccccccc|cccccccc|cccccccc|}
\hline 
Dataset & \multicolumn{8}{c|}{FID$\downarrow$} & \multicolumn{8}{c|}{KID$\downarrow$} & \multicolumn{8}{c|}{NIQE$\downarrow$} & \multicolumn{8}{c|}{$ID_{style}\downarrow$}& \multicolumn{8}{c|}{$ID_{content}\downarrow$}\tabularnewline
\hline 
\makecell[c]{Methods} & \makecell[c]{Style\\verse} & \makecell[c]{AAMS\\ \cite{aams}} & \makecell[c]{Avatar\\-net \cite{avatar}} &\makecell[c]{WCT\\ \cite{WCT}} & \makecell[c]{$WCT^{2}$\\ \cite{wct2}} & \makecell[c]{AdaIN\\ \cite{adain2017}} & \makecell[c]{SANet\\ \cite{sanet}} & \makecell[c]{DSTN\\ \cite{dstn}} & \makecell[c]{Style\\verse} & \makecell[c]{AAMS\\ \cite{aams}} & \makecell[c]{Avatar\\-net \cite{avatar}} &\makecell[c]{WCT\\ \cite{WCT}} & \makecell[c]{$WCT^{2}$\\ \cite{wct2}} & \makecell[c]{AdaIN\\ \cite{adain2017}} & \makecell[c]{SANet\\ \cite{sanet}} & \makecell[c]{DSTN\\ \cite{dstn}}&\makecell[c]{Style\\verse} & \makecell[c]{AAMS\\ \cite{aams}} & \makecell[c]{Avatar\\-net \cite{avatar}} &\makecell[c]{WCT\\ \cite{WCT}} & \makecell[c]{$WCT^{2}$\\ \cite{wct2}} & \makecell[c]{AdaIN\\ \cite{adain2017}} & \makecell[c]{SANet\\ \cite{sanet}} & \makecell[c]{DSTN\\ \cite{dstn}}&\makecell[c]{Style\\verse} & \makecell[c]{AAMS\\ \cite{aams}} & \makecell[c]{Avatar\\-net \cite{avatar}} &\makecell[c]{WCT\\ \cite{WCT}} & \makecell[c]{$WCT^{2}$\\ \cite{wct2}} & \makecell[c]{AdaIN\\ \cite{adain2017}} & \makecell[c]{SANet\\ \cite{sanet}} & \makecell[c]{DSTN\\ \cite{dstn}}&\makecell[c]{Style\\verse} & \makecell[c]{AAMS\\ \cite{aams}} & \makecell[c]{Avatar\\-net \cite{avatar}} &\makecell[c]{WCT\\ \cite{WCT}} & \makecell[c]{$WCT^{2}$\\ \cite{wct2}} & \makecell[c]{AdaIN\\ \cite{adain2017}} & \makecell[c]{SANet\\ \cite{sanet}} & \makecell[c]{DSTN\\ \cite{dstn}}\tabularnewline
 
\hline 
Daytime portrait & 181.84&	206.66&	199.01&	236.51&	202.6&	212.06&	\textbf{\textcolor{red}{178.8}}&	197.97&	6.8&	8.87&	8.84&	12.73&	7.61&	9.83&	\textbf{\textcolor{red}{6.24}}&	9.17&	7.75&	8.79&	7.52&	\textbf{\textcolor{red}{5.32}}&	6.15&	6.07&	7.09&	7.89&	0.86&	0.89&	0.84&	0.86&	0.88&	0.9&	\textbf{\textcolor{red}{0.84}}&	0.85&	0.44&	0.42&	0.38&	0.67&	\textbf{\textcolor{red}{0.08}}&	0.23&	0.36&	0.18\tabularnewline
3D cartoon & 146.56&	192.53&	168.59&	215.94&	174.21&	177.58&	\textbf{\textcolor{red}{141.8}}&	161.95&	3.46&	7.05&	5.77&	11.5&	5.79&	6.55&	\textbf{\textcolor{red}{2.91}}&	5.35&	7.65&	8.68&	7.7&	\textbf{\textcolor{red}{5.85}}&	6.3&	6.79&	7.21&	8&	0.67&	0.77&	0.73&	\textbf{\textcolor{red}{0.65}}&	0.82&	0.81&	0.7&	0.75&	0.48&	0.41&	0.39&	0.66&	\textbf{\textcolor{red}{0.1}}&	0.27&	0.37& 0.19\tabularnewline
Avatar & \textbf{\textcolor{red}{111.54}}&	228.62&	168.05&	258.8&	157.62&	148.63&	132.51&	136.45&	\textbf{\textcolor{red}{3.95}}&	13.15&	8.34&	20.11&	5.35&	4.97&	5.06&	4.74&	7.72&	8.87&	8.14&	\textbf{\textcolor{red}{6.27}}&	6.5&	7.1&	7.55&	8.14&	0.75&	0.81&	0.77&	\textbf{\textcolor{red}{0.73}}&	0.86&	0.83&	0.77&	0.8&	0.43&	0.47&	0.42&	0.7&	\textbf{\textcolor{red}{0.08}}&	0.29&	0.37&	0.23\tabularnewline
Anime & 147.19&	222.72&	200.24&	260.32&	194.47&	146.96&	\textbf{\textcolor{red}{123.35}}&	158.32&	6.35&	11.89&	11.03&	19.06&	9.24&	4.83&	\textbf{\textcolor{red}{3.67}}&	6.9&	7.63&	8.56&	7.41&	\textbf{\textcolor{red}{5.76}}&	6.37&	6.95&	7.07&	7.69&	0.62&	0.71&	0.71&	\textbf{\textcolor{red}{0.57}}&	0.82&	0.77&	0.71&	0.75&	0.55&	0.42&	0.45&	0.69&	\textbf{\textcolor{red}{0.07}}&	0.29&	0.35&	0.23\tabularnewline
Sculpture & 144.34&	223.91&	167.32&	236.88&	154.95&	150.68&	\textbf{\textcolor{red}{135.04}}&	150.20& 	5.43&	9.98&	6.56&	14.13&	4.89&	4.60& 	\textbf{\textcolor{red}{4.04}}&	5.17&	7.98&	9.37&	8.45&	\textbf{\textcolor{red}{6.72}}&	6.88&	7.66&	7.93&	8.42&	0.77&	0.8&	0.77&	\textbf{\textcolor{red}{0.74}}&	0.83&	0.83&	0.78&	0.79&	0.48&	0.46&	0.40& 	0.66&	\textbf{\textcolor{red}{0.06}}&	0.25&	0.29&	0.20 
\tabularnewline
Buddha statue & 138&	197.41&	182.94&	222.48&	155.85&	162.95&	\textbf{\textcolor{red}{131.26}}&	147.59&	6.6&	10.20& 	9.75&	13.42&	5.71&	7.09&	\textbf{\textcolor{red}{5.21}}&	6.57&	7.88&	9.1&	8.19&	\textbf{\textcolor{red}{6.19}}&	6.61&	7.47&	7.69&	8.14&	\textbf{\textcolor{red}{0.6}}&	0.78&	0.74&	0.67&	0.8&	0.8&	0.71&	0.75&	0.50& 	0.44&	0.41&	0.65&	\textbf{\textcolor{red}{0.13}}&	0.27&	0.34&	0.20 
\tabularnewline
Sketch & 203.24&	210.96&	193.4&	224.28&	224.77&	238.44&	168.91&	\textbf{\textcolor{red}{166.90}}& 	20.42&	18.84&	17.50& 	24.17&	19.16&	22.93&	15.88&	\textbf{\textcolor{red}{14.05}}&	7.35&	8.26&	7.23&	\textbf{\textcolor{red}{4.97}}&	5.86&	6.4&	6.26&	7.13&	\textbf{\textcolor{red}{0.68}}&	0.82&	0.76&	0.77&	0.87&	0.85&	0.77&	0.77&	0.52&	0.45&	0.35&	0.63&	\textbf{\textcolor{red}{0.11}}&	0.28&	0.34&	0.21
\tabularnewline
Ancient portrait & 152.5&	214.05&	191.18&	271.51&	167.61&	158.89&	\textbf{\textcolor{red}{137.27}}&	146.51&	5.01&	7.18&	7.29&	15.36&	3.77&	3.79&	\textbf{\textcolor{red}{3.16}}&	3.64&	7.98&	9.42&	8.42&	\textbf{\textcolor{red}{6.56}}&	6.9&	7.68&	8.02&	8.47&	\textbf{\textcolor{red}{0.65}}&	0.78&	0.74&	0.69&	0.81&	0.8&	0.73&	0.75&	0.51&	0.47&	0.42&	0.67&	\textbf{\textcolor{red}{0.09}}&	0.29&	0.32&	0.20 
\tabularnewline
NIR & 207.47&	219.34&	246.66&	276.48&	214.50& 	238.28&	\textbf{\textcolor{red}{170.06}}&	183.89&	17.2&	18.53&	24.05&	29.75&	15.40& 	20.98&	14.34&	\textbf{\textcolor{red}{14.30}}& 	7.69&	8.57&	8.27&	6.38&	\textbf{\textcolor{red}{6.37}}&	6.57&	7.42&	7.9&	\textbf{\textcolor{red}{0.68}}&	0.86&	0.8&	0.82&	0.87&	0.88&	0.82&	0.82&	0.62&	0.53&	0.47&	0.68&	\textbf{\textcolor{red}{0.1}}&	0.35&	0.30& 	0.22
\tabularnewline
Beijing Opera & \textbf{\textcolor{red}{125.39}}&	196.62&	158.35&	197.45&	159.47&	159.88&	131.02&	145.55&	5.18&	10.27&	7.77&	12.63&	5.88&	6.63&	\textbf{\textcolor{red}{4.84}}&	6.08&	7.76&	8.86&	7.81&	\textbf{\textcolor{red}{6.22}}&	6.31&	7.1&	7.4&	8.06&	\textbf{\textcolor{red}{0.67}}&	0.81&	0.78&	0.76&	0.84&	0.84&	0.75&	0.78&	0.54&	0.42&	0.38&	0.65&	\textbf{\textcolor{red}{0.09}}&	0.26&	0.35&	0.19
\tabularnewline
Exaggerated drawing & \textbf{\textcolor{red}{129.58}}&	183.62&	169.96&	219.09&	151.84&	151.43&	140.05&	139.37&	4.17&	7.77&	8.45&	14.76&	\textbf{\textcolor{red}{4.14}}&	5.60& 	4.41&	4.55&	7.52&	8.4&	7.35&	\textbf{\textcolor{red}{5.47}}&	6.15&	6.02&	6.93&	7.75&	0.83&	0.81&	0.78&	\textbf{\textcolor{red}{0.74}}&	0.84&	0.84&	0.77&	0.80& 	0.43&	0.41&	0.38&	0.66&	\textbf{\textcolor{red}{0.09}}&	0.24&	0.39&	0.21
\tabularnewline
Oil painting & 157.89&	210.51&	181.21&	228.02&	156.56&	149.63&	\textbf{\textcolor{red}{126.09}}&	144.45&	7.36&	13.5&	11.02&	15.81&	6.00& 	5.90&	\textbf{\textcolor{red}{5.17}}&	6.60& 	7.85&	9.54&	8.66&	\textbf{\textcolor{red}{5.86}}&	6.71&	6.46&	8.08&	8.79&	\textbf{\textcolor{red}{0.78}}&	0.84&	0.81&	0.81&	0.86&	0.88&	0.82&	0.82&	0.49&	0.49&	0.42&	0.68&	\textbf{\textcolor{red}{0.07}}&	0.26&	0.32&	0.20 
\tabularnewline
Night portrait & 148.83&	210.24&	180.39&	240.09&	190.17&	172.15&	\textbf{\textcolor{red}{140.14}}&	158.72&	4.28&	9.03&	7.85&	13.77&	8.22&	6.01&	\textbf{\textcolor{red}{3.55}}&	5.30& 	7.93&	9.33&	8.64&	6.89&	\textbf{\textcolor{red}{6.79}}&	7.88&	8.17&	8.33&	0.83&	0.86&	0.83&	0.84&	0.85&	0.88&	\textbf{\textcolor{red}{0.83}}&	0.84&	0.48&	0.48&	0.38&	0.66&	\textbf{\textcolor{red}{0.1}}&	0.31&	0.32&	0.18
\tabularnewline
\hline 
Mean & \textbf{153.41}&	209.01&	185.17&	237.52&	177.27&	174.42&	\textbf{\textcolor{red}{142.79}}&	156.75&	\textbf{7.40}& 	11.25&	10.32&	16.71&	7.78&	8.43&	\textbf{\textcolor{red}{6.03}}&	7.11&	\textbf{7.74}&	8.9&	7.98&	\textbf{\textcolor{red}{6.03}}&	6.45&	6.93&	7.44&	8.05&	\textbf{\textcolor{red}{0.72}}&	0.81&	0.77&	0.74&	0.84&	0.84&	0.77&	0.79&	\textbf{0.49}&	0.45&	0.40& 	0.66&	\textbf{\textcolor{red}{0.09}}&	0.27&	0.34&	0.20 \tabularnewline
\hline
\end{tabular}
}
\vspace{-10.pt}
\label{tab: st}
\end{table*}
\begin{table*}[h]
\caption{Quantitative evaluation of ablation study with FID, KID$\times100$,NIQE, $ID_{style}$ and $ID_{content}$. The red values mean the top-1 scores for each row, among Styleverse (VGG\_LightCNN), VGG\_$Arc^{2}$, $Arc^{1}$\_LightCNN, $Arc^{1}$\_$Arc^{2}$, $Arc^{1}$ and w/ $\mathcal{L}_{CLIP}$. VGG\_Arc has the best $ID_{content}$ score but the worst $ID_{style}$ score, which indicates LightCNN is better at extracting distinctive PSU style than Arcface. Although $Arc^{1}$ has the best FID  and $ID_{style}$ scores, but it inclines to identity swapping rather than identity stylization.}
\vspace{-10.pt}
\scalebox{0.302}{
\begin{tabular}{|c|cccccc|cccccc|cccccc|cccccc|cccccc|}
\hline 
Dataset & \multicolumn{6}{c|}{FID$\downarrow$} & \multicolumn{6}{c|}{KID$\downarrow$} & \multicolumn{6}{c|}{NIQE$\downarrow$} & \multicolumn{6}{c|}{$ID_{style}\downarrow$}& \multicolumn{6}{c|}{$ID_{content}\downarrow$}\tabularnewline
\hline 
\makecell[c]{Methods} & \makecell[c]{Styleverse} & \makecell[c]{VGG\_$Arc^{2}$} & \makecell[c]{$Arc^{1}$\_\\LightCNN} &\makecell[c]{$Arc^{1}$\_$Arc^{2}$} & \makecell[c]{$Arc^{1}$} & \makecell[c]{w/ $\mathcal{L}_{CLIP}$} & \makecell[c]{Styleverse} & \makecell[c]{VGG\_$Arc^{2}$} & \makecell[c]{$Arc^{1}$\_\\LightCNN} &\makecell[c]{$Arc^{1}$\_$Arc^{2}$} & \makecell[c]{$Arc^{1}$} & \makecell[c]{w/ $\mathcal{L}_{CLIP}$} & \makecell[c]{Styleverse} & \makecell[c]{VGG\_$Arc^{2}$} & \makecell[c]{$Arc^{1}$\_\\LightCNN} &\makecell[c]{$Arc^{1}$\_$Arc^{2}$} & \makecell[c]{$Arc^{1}$} & \makecell[c]{w/ $\mathcal{L}_{CLIP}$} & \makecell[c]{Styleverse} & \makecell[c]{VGG\_$Arc^{2}$} & \makecell[c]{$Arc^{1}$\_\\LightCNN} &\makecell[c]{$Arc^{1}$\_$Arc^{2}$} & \makecell[c]{$Arc^{1}$} & \makecell[c]{w/ $\mathcal{L}_{CLIP}$} & \makecell[c]{Styleverse} & \makecell[c]{VGG\_$Arc^{2}$} & \makecell[c]{$Arc^{1}$\_\\LightCNN} &\makecell[c]{$Arc^{1}$\_$Arc^{2}$} & \makecell[c]{$Arc^{1}$} & \makecell[c]{w/ $\mathcal{L}_{CLIP}$}\tabularnewline
 
\hline 
Daytime portrait & 181.84&	184.28&	175.24&	172.45&	\textbf{\textcolor{red}{169.81}}&	189.95&	6.8&	6.72&	6.2&	5.62&	\textbf{\textcolor{red}{5.51}}&	8.15&	\textbf{\textcolor{red}{7.75}}&	7.85&	7.78&	7.86&	7.9&	7.96&	0.86&	0.87&	0.82&	0.79&	\textbf{\textcolor{red}{0.49}}&	0.86&	0.44&	\textbf{\textcolor{red}{0.32}}&	0.46&	0.39&	0.61&	0.44
\tabularnewline
3D cartoon & 146.56&	146.5&	146.36&	145.16&	\textbf{\textcolor{red}{142.98}}&	147.23&	3.46&	3.46&	3.43&	3.32&	\textbf{\textcolor{red}{3.02}}&	3.82&	7.65&	7.79&	\textbf{\textcolor{red}{7.64}}&	7.66&	7.78&	7.81&	0.67&	0.72&	0.65&	0.65&	\textbf{\textcolor{red}{0.38}}&	0.68&	0.48&	\textbf{\textcolor{red}{0.37}}&	0.47&	0.4&	0.61&	0.45
\tabularnewline
Avatar & 111.54&	123.64&	\textbf{\textcolor{red}{109.05}}&	125.56&	113.81&	112.75&	3.95&	4.22&	\textbf{\textcolor{red}{3.56}}&	4.23&	3.68&	4&	\textbf{\textcolor{red}{7.72}}&	7.8&	7.72&	7.76&	7.89&	7.87&	0.75&	0.81&	0.71&	0.74&	\textbf{\textcolor{red}{0.33}}&	0.77&	0.43&	\textbf{\textcolor{red}{0.31}}&	0.44&	0.36&	0.65&	0.38
\tabularnewline
Anime & 147.19&	148.01&	146.71&	153.93&	\textbf{\textcolor{red}{145.5}}&	148.56&	6.35&	\textbf{\textcolor{red}{6.11}}&	6.47&	7.03&	6.51&	6.42&	7.63&	7.72&	\textbf{\textcolor{red}{7.61}}&	7.73&	7.72&	7.8&	0.62&	0.72&	0.59&	0.69&	\textbf{\textcolor{red}{0.42}}&	0.61&	0.55&	\textbf{\textcolor{red}{0.42}}&	0.56&	0.44&	0.66&	0.54
\tabularnewline
Sculpture & 144.34&	158.84&	143.98&	146.17&	\textbf{\textcolor{red}{141.78}}&	145.05&	5.43&	6.58&	5.21&	5.23&	\textbf{\textcolor{red}{4.78}}&	5.43&	7.98&	8.11&	\textbf{\textcolor{red}{7.97}}&	7.99&	8.05&	8.11&	0.77&	0.80& 	0.73&	0.72&	\textbf{\textcolor{red}{0.44}}&	0.77&	0.48&	\textbf{\textcolor{red}{0.34}}&	0.49&	0.42&	0.62&	0.45
\tabularnewline
Buddha statue & 138&	144.35&	136.74&	145.17&	\textbf{\textcolor{red}{131.45}}&	141.19&	6.6&	6.82&	6.41&	6.42&	\textbf{\textcolor{red}{5.62}}&	6.84&	\textbf{\textcolor{red}{7.88}}&	8.02&	7.93&	7.98&	8&	8.08&	0.6&	0.66&	0.58&	0.60& 	\textbf{\textcolor{red}{0.35}}&	0.6&	0.50& 	\textbf{\textcolor{red}{0.38}}&	0.50& 	0.45&	0.64&	0.48
\tabularnewline
Sketch & 203.24&	205.34&	199.73&	203.81&	\textbf{\textcolor{red}{195.06}}&	202.9&	20.42&	20.25&	19.81&	18.84&	\textbf{\textcolor{red}{18.14}}&	20.69&	\textbf{\textcolor{red}{7.35}}&	7.6&	7.59&	7.58&	7.55&	7.42&	0.68&	0.73&	0.66&	0.68&	\textbf{\textcolor{red}{0.42}}&	0.7&	0.52&	\textbf{\textcolor{red}{0.4}}&	0.54&	0.46&	0.65&	0.5
\tabularnewline
Ancient portrait & 152.5&	150.47&	150.56&	148.29&	\textbf{\textcolor{red}{147.75}}&	157.49&	5.01&	4.63&	4.77&	\textbf{\textcolor{red}{3.63}}&	3.92&	5.64&	\textbf{\textcolor{red}{7.98}}&	8.09&	7.98&	8.02&	8.09&	8.14&	0.65&	0.69&	0.63&	0.63&	\textbf{\textcolor{red}{0.37}}&	0.67&	0.51&	\textbf{\textcolor{red}{0.4}}&	0.52&	0.43&	0.63&	0.49
\tabularnewline
NIR & 207.47&	203.06&	207.32&	204.81&	\textbf{\textcolor{red}{200.65}}&	207.92&	17.2&	16.99&	17.65&	17.47&	\textbf{\textcolor{red}{16.75}}&	18.11&	\textbf{\textcolor{red}{7.69}}&	7.86&	7.74&	7.78&	7.97&	7.87&	0.68&	0.74&	0.59&	0.65&	\textbf{\textcolor{red}{0.36}}&	0.71&	0.62&	\textbf{\textcolor{red}{0.48}}&	0.67&	0.50& 	0.71&	0.6
\tabularnewline
Beijing Opera & 125.39&	125.77&	122.66&	123.57&	\textbf{\textcolor{red}{120.34}}&	126.87&	5.18&	4.72&	4.82&	\textbf{\textcolor{red}{4.57}}&	4.74&	5.39&	\textbf{\textcolor{red}{7.76}}&	7.88&	7.77&	7.86&	7.93&	7.87&	0.67&	0.71&	0.62&	0.65&	\textbf{\textcolor{red}{0.31}}&	0.68&	0.54&	\textbf{\textcolor{red}{0.43}}&	0.56&	0.48&	0.67&	0.53
\tabularnewline
Exaggerated drawing &129.58&	132.38&	125.16&	130.15&	\textbf{\textcolor{red}{122.84}}&	130.3&	4.17&	4&	3.71&	3.81&	\textbf{\textcolor{red}{3.16}}&	4.33&	\textbf{\textcolor{red}{7.52}}&	7.7&	7.54&	7.65&	7.79&	7.81&	0.83&	0.83&	0.80& 	0.78&	\textbf{\textcolor{red}{0.49}}&	0.82&	0.43&	\textbf{\textcolor{red}{0.29}}&	0.43&	0.32&	0.57&	0.41
\tabularnewline
Oil painting & 157.89&	155.14&	156.87&	\textbf{\textcolor{red}{154.75}}&	158.66&	160.1&	7.36&	7.06&	7.45&	\textbf{\textcolor{red}{6.91}}&	7.11&	7.55&	7.85&	7.96&	\textbf{\textcolor{red}{7.81}}&	7.89&	7.93&	7.96&	0.78&	0.80& 	0.73&	0.74&	\textbf{\textcolor{red}{0.43}}&	0.77&	0.49&	\textbf{\textcolor{red}{0.38}}&	0.50 &	0.41&	0.63&	0.49
\tabularnewline
Night portrait & 148.83&	146.09&	143.86&	151.54&	\textbf{\textcolor{red}{139.73}}&	148.05&	4.28&	4.12&	\textbf{\textcolor{red}{3.49}}&	4.69&	3.6&	4.44&	7.93&	7.98&	7.95&	\textbf{\textcolor{red}{7.89}}&	8.04&	8.01&	0.83&	0.85&	0.78&	0.77&	\textbf{\textcolor{red}{0.45}}&	0.83&	0.48&	\textbf{\textcolor{red}{0.35}}&	0.48&	0.40& 	0.61&	0.45
\tabularnewline
\hline 
Mean & \textbf{153.41}&	155.68&	151.09&	154.25&	\textbf{\textcolor{red}{148.49}}&	155.25&	\textbf{7.40} &	7.36&	7.15&	7.06&	\textbf{\textcolor{red}{6.65}}&	7.75&	\textbf{\textcolor{red}{7.74}}&	7.87&	7.77&	7.82&	7.89&	7.90 &	\textbf{0.72}&	0.76&	0.68&	0.70& 	\textbf{\textcolor{red}{0.4}}&	0.73&	\textbf{0.49}&	\textbf{\textcolor{red}{0.37}}&	0.51&	0.42&	0.63&	0.47
\tabularnewline
\hline
\end{tabular}
}
\vspace{-16.pt}
\label{tab: ablation}
\end{table*}
\paragraph{Contextual loss weight} 
As shown in Figure \ref{fig:weight}, the weight of $\mathcal{L}_{CCX}$ controls the degree of content preservation, \ie, higher $\lambda_{CCX}$ helps to maintain more source content, but prevents sufficient facial style transfer, \eg, $CCX_{0.5}$. Styleverse utilizes $z_{DIV}$ based on $CCX_{0.5}$, to synthesize more natural IDS results, \eg, the makeup and skin color of the female role of Beijing Opera are reasonably swapped on the diverse contents. Our IDS results are highly topology-aware and maintain the content identity.

\paragraph{IDS diversity} 
As shown in Figure \ref{fig:div}, Styleverse is competent to synthesizing assigned IDS productions given specific target reference images, \eg, the skin color and facial component features are controllably adapted to the source contents of diverse PSUs. Nevertheless, it is difficult for w/o $z_{DIV}$ to handle the reference-based IDS.

\paragraph{Quantitative evaluation} 
As shown in Table \ref{tab: st}, we establish the first benchmark of IDS task, compared with state-of-the-art style transfer methods. Specifically, we choose 10 content images of each PSU as the content set, and implement IDS based on diverse heterogeneous domains. The qualitative results are shown in Figure \ref{fig:very_big}. We show the quantitative values considering several image fidelity evaluation metrics, \ie, FID \cite{FID}, KID \cite{KID} and NIQE \cite{NIQE}. Specifically, KID \cite{KID} is the Kernel Inception Distance using the squared Maximum Mean Discrepancy (MMD) with a polynomial kernel, which is used in \cite{ugatit,nicegan}. NIQE \cite{NIQE} is a non-reference blind image quality assessment for grayscale images. FID and KID are measured based on the same PSU style ground-truth for IDS results of different methods. WCT \cite{WCT} has the best NIQE scores, but with worse topology awareness, as shown in Figure \ref{fig:very_big}.
\begin{table*}[h]
\caption{Quantitative evaluation of Styleverse on FS13 with FID, KID$\times100$ and NIQE. The black, red and blue values mean the top-1, top-2 and top-3 scores for each column. Note that the first raw means the style and the first column means the content. FID and KID are measured based on the same PSU style ground-truth for IDS results with different contents.}
\vspace{-10.pt}
\scalebox{0.31}{
\begin{tabular}{|c|ccc|ccc|ccc|ccc|ccc|ccc|ccc|ccc|ccc|ccc|ccc|ccc|ccc|}
\hline 
Dataset & \multicolumn{3}{c|}{Daytime portrait} & \multicolumn{3}{c|}{3D cartoon} & \multicolumn{3}{c|}{Avatar} & \multicolumn{3}{c|}{Anime}& \multicolumn{3}{c|}{Sculpture} & \multicolumn{3}{c|}{Buddha statue} & \multicolumn{3}{c|}{Sketch} & \multicolumn{3}{c|}{Ancient portrait}& \multicolumn{3}{c|}{NIR} & \multicolumn{3}{c|}{Beijing Opera} & \multicolumn{3}{c|}{Exaggerated drawing} & \multicolumn{3}{c|}{Oil painting}& \multicolumn{3}{c|}{Night portrait}\tabularnewline
\hline 
\makecell[c]{Methods} & \makecell[c]{FID$\downarrow$} & \makecell[c]{KID$\downarrow$} & NIQE$\downarrow$ &\makecell[c]{FID$\downarrow$} & \makecell[c]{KID$\downarrow$} & NIQE$\downarrow$ & \makecell[c]{FID$\downarrow$} & \makecell[c]{KID$ \downarrow$} & NIQE$\downarrow$ &\makecell[c]{FID$\downarrow$} & \makecell[c]{KID$ \downarrow$} & NIQE$\downarrow$ & \makecell[c]{FID$\downarrow$} & \makecell[c]{KID$ \downarrow$} & NIQE$\downarrow$ &\makecell[c]{FID$\downarrow$} & \makecell[c]{KID$ \downarrow$} & NIQE$\downarrow$ & \makecell[c]{FID$\downarrow$} & \makecell[c]{KID$ \downarrow$} & NIQE$\downarrow$ &\makecell[c]{FID$\downarrow$} & \makecell[c]{KID$ \downarrow$} & NIQE$\downarrow$ & \makecell[c]{FID$\downarrow$} & \makecell[c]{KID$ \downarrow$} & NIQE$\downarrow$ &\makecell[c]{FID$\downarrow$} & \makecell[c]{KID$\downarrow$} & NIQE$\downarrow$ & \makecell[c]{FID$\downarrow$} & \makecell[c]{KID$ \downarrow$} & NIQE$\downarrow$ & \makecell[c]{FID$\downarrow$} & \makecell[c]{KID$ \downarrow$} & NIQE$\downarrow$ & \makecell[c]{FID$\downarrow$} & \makecell[c]{KID$ \downarrow$} & NIQE$\downarrow$\tabularnewline
 
\hline 
Daytime portrait & \textbf{76.91}&	\textbf{1.41}&	7.93&	170.97&	5.71&	7.59&	150.46&	7.43&	7.61&	191.28&	11.24&	\textbf{\textcolor{blue}{7.40}}&	203.64&	11.37&	7.85&	175.71&	10.65&	7.59&	210.47&	\textbf{\textcolor{blue}{20.32}}&	7.08&	191.32&	9.34&	7.81&	233.84&	\textbf{\textcolor{blue}{19.97}}&	\textbf{\textcolor{blue}{7.60}}&	154.18&	7.78&	7.54&	174.97&	8.18&	7.62&	169.30&	\textbf{\textcolor{red}{7.96}}&	7.81&	162.28&	\textbf{\textcolor{red}{4.89}}&	\textbf{\textcolor{blue}{7.72}}\tabularnewline
3D cartoon & 207.60&	\textbf{\textcolor{blue}{9.82}}&	7.72&	\textbf{82.92}&	\textbf{1.48}&	7.99&	126.87&	5.71&	7.65&	160.01&	8.49&	7.59&	187.21&	10.95&	7.92&	177.84&	11.82&	7.83&	241.37&	27.02&	\textbf{\textcolor{red}{7.17}}&	193.72&	9.95&	7.96&	262.27&	26.59&	7.69&	150.32&	7.83&	7.69&	172.87&	9.18&	7.53&	199.73&	12.32&	7.84&	\textbf{\textcolor{blue}{161.12}}&	\textbf{\textcolor{blue}{5.10}}&	7.94\tabularnewline
Avatar & 263.70&	17.53&	7.69&	184.36&	8.97&	7.58&	\textbf{63.14}&	\textbf{1.66}&	8.02&	159.67&	8.88&	7.63&	178.81&	10.94&	7.94&	182.07&	13.42&	7.96&	254.34&	30.45&	7.31&	187.86&	9.87&	7.89&	264.53&	28.18&	7.73&	\textbf{\textcolor{blue}{127.16}}&	\textbf{\textcolor{blue}{6.01}}&	7.74&	156.38&	8.03&	7.46&	255.84&	21.52&	7.78&	195.59&	11.25&	7.91\tabularnewline
Anime & 225.40&	13.69&	7.54&	160.09&	6.74&	7.63&	128.02&	7.25&	\textbf{\textcolor{blue}{7.56}}&	\textbf{65.10}&	\textbf{1.19}&	7.77&	170.41&	9.80&	7.83&	152.33&	10.13&	7.80&	257.35&	30.84&	7.28&	\textbf{\textcolor{blue}{162.83}}&	\textbf{\textcolor{blue}{6.96}}&	7.81&	250.91&	26.66&	7.63&	128.78&	6.53&	7.69&	151.08&	8.29&	7.48&	216.15&	16.05&	7.72&	186.67&	10.39&	7.74\tabularnewline
Sculpture & 223.56&	11.58&	7.86&	154.71&	5.09&	7.80&	\textbf{\textcolor{red}{115.19}}&	\textbf{\textcolor{blue}{4.85}}&	8.03&	170.84&	10.26&	7.83&	\textbf{73.46}&	\textbf{1.23}&	8.71&	\textbf{\textcolor{red}{139.43}}&	\textbf{\textcolor{red}{7.15}}&	8.17&	226.69&	23.84&	7.45&	169.80&	\textbf{\textcolor{red}{6.84}}&	8.20&	252.46&	25.39&	7.91&	137.45&	6.64&	7.99&	153.12&	\textbf{\textcolor{blue}{7.29}}&	7.73&	181.99&	10.85&	8.00&	178.28&	8.59&	8.18\tabularnewline
Buddha statue & 211.40&	11.68&	7.83&	153.45&	5.55&	7.87&	129.8&	7.24&	7.96&	164.00&	10.22&	7.84&	\textbf{\textcolor{red}{140.94}}&	\textbf{\textcolor{red}{5.98}}&	8.17&	\textbf{58.2}&	\textbf{1.46}&	8.37&	223.68&	25.48&	7.24&	167.54&	7.43&	8.32&	254.57&	27.00& 	7.91&	133.10&	7.43&	8.05&	171.83&	10.78&	7.64&	\textbf{\textcolor{red}{166.21}}&	\textbf{\textcolor{blue}{10.19}}&	7.95&	163&	7.14&	8.20\tabularnewline
Sketch & 219.82&	15.84&	\textbf{7.25}&	178.27&	11.10& 	7.31&	132.68&	8.81&	\textbf{7.22}&	179.87&	14.01&	\textbf{7.11}&	179.69&	14.53&	\textbf{7.56}&	\textbf{\textcolor{blue}{140.50}}&	10.96&	\textbf{7.39}&	\textbf{83.01}&	\textbf{8.76}&	7.60&	177.69&	11.60& 	\textbf{7.62}&	\textbf{\textcolor{red}{165.14}}&	\textbf{\textcolor{red}{19.06}}&	\textbf{7.35}&	176.47&	15.81&	\textbf{7.37}&	\textbf{\textcolor{red}{131.37}}&	7.46&	\textbf{7.06}&	\textbf{\textcolor{blue}{167.28}}&	13.36&	\textbf{7.47}&	170.67&	10.49&	\textbf{\textcolor{red}{7.53}}\tabularnewline
Ancient portrait & 216.32&	10.51&	7.88&	162.18&	5.32&	7.87&	122.29&	4.91&	7.85&	164.23&	\textbf{\textcolor{blue}{8.11}}&	7.90&	163.57&	\textbf{\textcolor{blue}{7.31}}&	8.18&	145.71&	\textbf{\textcolor{blue}{7.64}}&	8.19&	225.53&	24.05&	7.53&	\textbf{79.55}&	\textbf{1.15}&	8.53&	253.91&	24.90& 	7.93&	\textbf{\textcolor{red}{120.77}}&	\textbf{\textcolor{red}{4.29}}&	8.05&	154.01&	\textbf{\textcolor{red}{6.94}}&	7.64&	185.13&	10.60& 	8.09&	183.46&	8.43&	8.18\tabularnewline
NIR & 209.86&	12.80& 	\textbf{\textcolor{red}{7.47}}&	185.76&	11.73&	\textbf{7.31}&	123.68&	7.67&	7.58&	177.61&	13.33&	\textbf{\textcolor{red}{7.26}}&	168.39&	11.48&	\textbf{\textcolor{red}{7.71}}&	148.56&	10.98&	\textbf{\textcolor{red}{7.59}}&	\textbf{\textcolor{red}{167.89}}&	\textbf{\textcolor{red}{19.75}}&	7.23&	178.91&	10.64&	\textbf{\textcolor{red}{7.66}}&	\textbf{69.68}&	\textbf{5.40}&	7.74&	155.41&	12.73&	\textbf{\textcolor{red}{7.53}}&	\textbf{\textcolor{blue}{140.56}}&	8.25&	\textbf{\textcolor{red}{7.28}}&	171.40&	12.57&	\textbf{\textcolor{red}{7.63}}&	165.11&	9.79&	\textbf{7.52}\tabularnewline
Beijing Opera & 230.91&	13.09&	\textbf{\textcolor{blue}{7.48}}&	168.96&	7.23&	\textbf{\textcolor{red}{7.48}}&	123.64&	5.97&	7.56&	\textbf{\textcolor{blue}{155.17}}&	8.40& 	7.44&	182.2&	10.86&	7.74&	155.98&	10.30& 	7.78&	265.82&	31.55&	\textbf{\textcolor{blue}{7.22}}&	\textbf{\textcolor{red}{161.34}}&	7.14&	7.75&	258.77&	27.49&	\textbf{\textcolor{red}{7.45}}&	\textbf{49.49}&	\textbf{0.80}&	7.79&	171.28&	10.25&	\textbf{\textcolor{blue}{7.33}}&	220.93&	17.10& 	7.63&	193.75&	11.18&	7.79\tabularnewline
Exaggerated drawing & 226.06&	13.31&	7.73&	176.39&	8.47&	\textbf{\textcolor{blue}{7.53}}&	\textbf{\textcolor{blue}{115.81}}&	\textbf{\textcolor{red}{4.84}}&	\textbf{\textcolor{red}{7.35}}&	\textbf{\textcolor{red}{150.63}}&	\textbf{\textcolor{red}{7.77}}&	7.50&	176.97&	10.86&	\textbf{\textcolor{blue}{7.72}}&	175.86&	12.86&	\textbf{\textcolor{blue}{7.70}}&	218.07&	25.10& 	\textbf{7.08}&	170.89&	8.56&	\textbf{\textcolor{blue}{7.67}}&	243.64&	25.44&	7.67&	143.76&	8.32&	\textbf{\textcolor{blue}{7.54}}&	\textbf{64.35}&	\textbf{1.72}&	7.61&	212.09&	15.71&	\textbf{\textcolor{blue}{7.64}}&	180.37&	9.72&	7.74\tabularnewline
Oil painting & \textbf{\textcolor{red}{174.87}}&	\textbf{\textcolor{red}{6.75}}&	8.01&	\textbf{\textcolor{red}{146.32}}&	\textbf{\textcolor{red}{4.57}}&	7.77&	130.56&	7.50& 	7.81&	187.57&	12.92&	7.58&	\textbf{\textcolor{blue}{157.72}}&	8.48&	8.11&	143.45&	9.00& 	7.84&	\textbf{\textcolor{blue}{189.89}}&	21.70& 	7.35&	172.17&	7.76&	8.20&	\textbf{\textcolor{blue}{213.61}}&	20.78&	7.69&	153.53&	9.96&	7.77&	151.72&	7.68&	7.67&	\textbf{69.41}&	\textbf{2.38}&	8.49&	\textbf{\textcolor{red}{153.25}}&	5.58&	7.89\tabularnewline
Night portrait & \textbf{\textcolor{blue}{205.71}}&	10.06&	7.92&	\textbf{\textcolor{blue}{151.72}}&	\textbf{\textcolor{blue}{4.82}}&	7.89&	130.63&	5.89&	8.10&	184.09&	11.81&	7.87&	193.53&	12.42&	8.25&	181.55&	12.64&	8.18&	238.36&	26.67&	7.39&	197.60&	11.53&	8.31&	241.27&	23.82&	7.94&	158.44&	9.49&	8.13&	174.48&	11.15&	7.74&	194.14&	12.25&	7.98&	\textbf{66.52}&	\textbf{0.76}&	8.49\tabularnewline
\hline 
\end{tabular}
}
\vspace{-16.pt}
\label{tab: nb}
\end{table*}

\begin{figure}[ht]
\begin{center}
\includegraphics[width=1\linewidth]{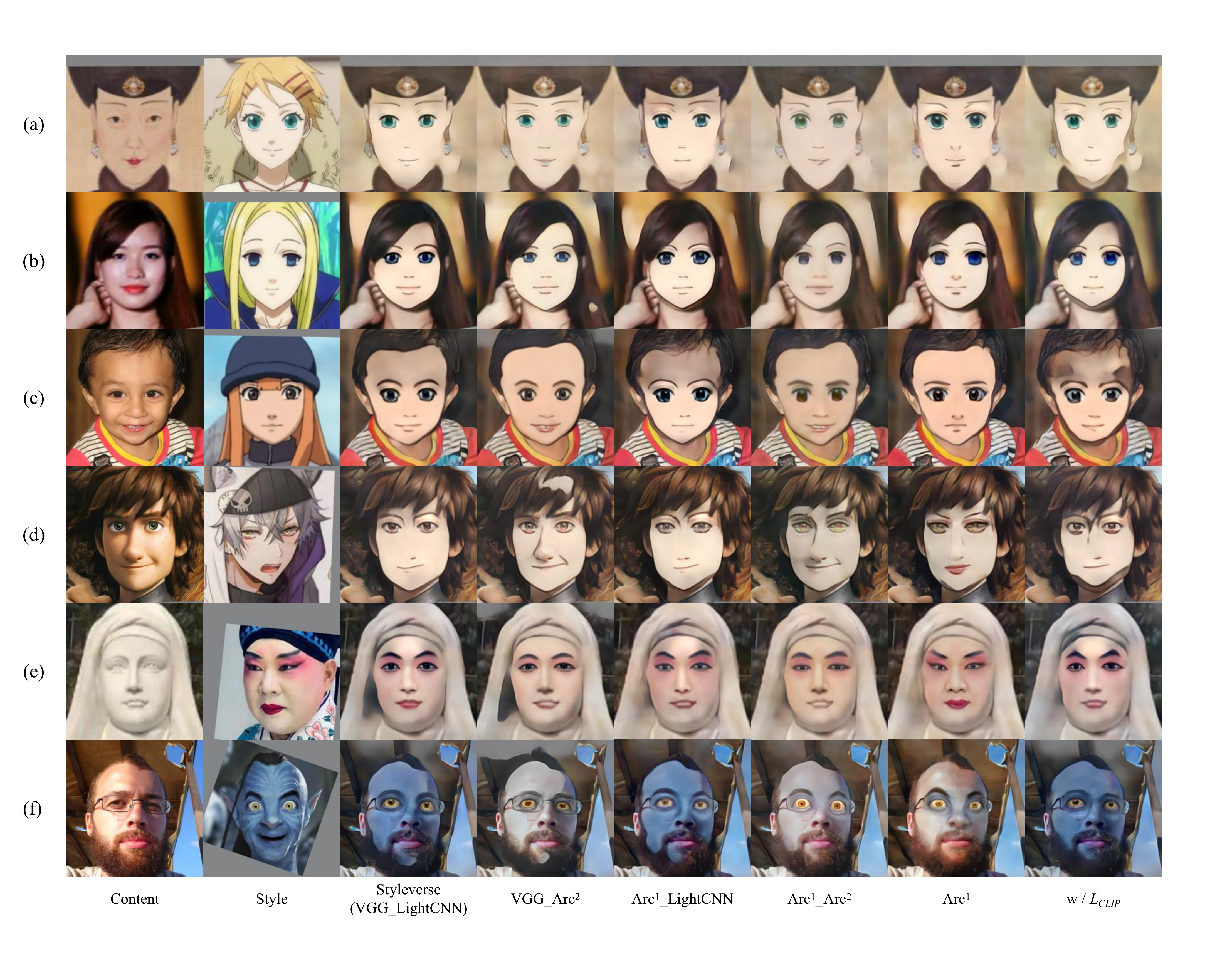}
\end{center}
\vspace{-13pt}
   \caption{Visual IDS comparison of Styleverse and the variants based on Arcface \cite{arcface2019}. $VGG\_Arc^{2}$ has insufficient IDS results, \eg, mouth in (c), and some artifacts, \eg, hair in (d), background in (e-f). $Arc^{1}\_LightCNN$ has excessive IDS compared with Styleverse, \eg, the face contours in (b) and (c).  $Arc^{1}\_Arc^{2}$ has blurred IDS results, \eg, (b, c, e), and worse NIQE scores as shown in Table \ref{tab: ablation}. $Arc^{1}$ has excessive identity swapping and ignores the expression of contents, \eg, (b, c, e). While w/ $\mathcal{L}_{CLIP}$ conducts lower-fidelity IDS (a, c, d). LightCNN is more competent to extracting heterogeneous-domain identity features, as shown in Figure \ref{fig:gbp}.}
\label{fig:arc}
\vspace{-10pt}
\end{figure}

As for the identity consistency, $WCT^{2}$ \cite{wct2} has the best $ID_{content}$ score, but with worse $ID_{style}$. Our Styleverse gets most of the best $ID_{style}$ scores, and has a highly topology-aware IDS performance, as shown in Figure \ref{fig:begin} and \ref{fig:very_big}.
\paragraph{Ablation study}
We conduct the ablation study as shown in Table \ref{tab: ablation} and Figure \ref{fig:arc}. Specifically, we compare different variants with Conv-BN-ReLU based Arcface as the texture encoder or distinctive style encoder. $Arc^{1}$ means the original instance-level identity recognition network, and $Arc^{2}$ is the pretrained PSU recognition network based on FS13. $VGG\_Arc^{2}$ has the best $ID_{content}$ score but the worst $ID_{style}$ score, which indicates LightCNN is better at extracting distinctive PSU styles than Arcface. Although only $w/ Arc^{1}$ has the best FID  and $ID_{style}$ scores, but it tends to conduct identity swapping rather than identity stylization. Furthermore, we combine a Contrastive Language-Image Pretraining (CLIP) \cite{clip} perceptual loss to constrain Styleverse in response to the heterogeneous-domain text prompt. We find the pretrained LightCNN or Arcface is better at guiding IDS than CLIP. As for the qualitative comparison in Figure \ref{fig:arc}, our Styleverse conducts more controllable IDS.
\begin{figure}[t]
\begin{center}
\includegraphics[width=0.9\linewidth]{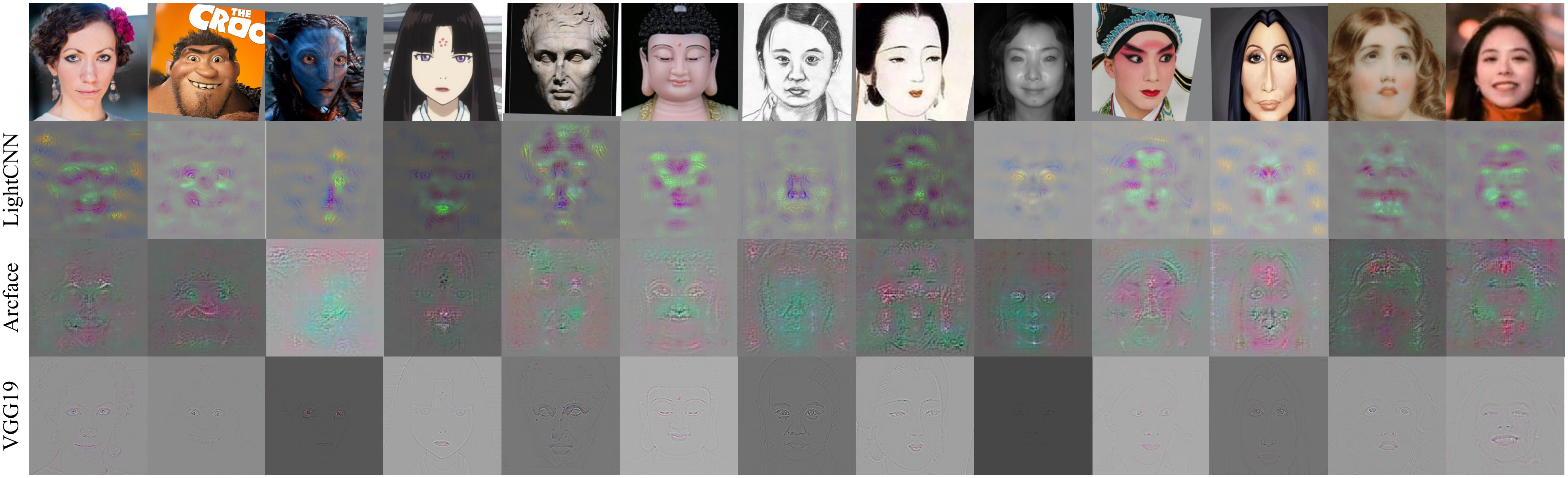}
\end{center}
\vspace{-13pt}
   \caption{Guided backpropagation (GBP) \cite{gbp} of LightCNN, Arcface and VGG19. GBP maps of LightCNN focus more on the topological face area.}
\label{fig:gbp}
\vspace{-10pt}
\end{figure}

\begin{figure}[t]
\begin{center}
\includegraphics[width=0.6\linewidth]{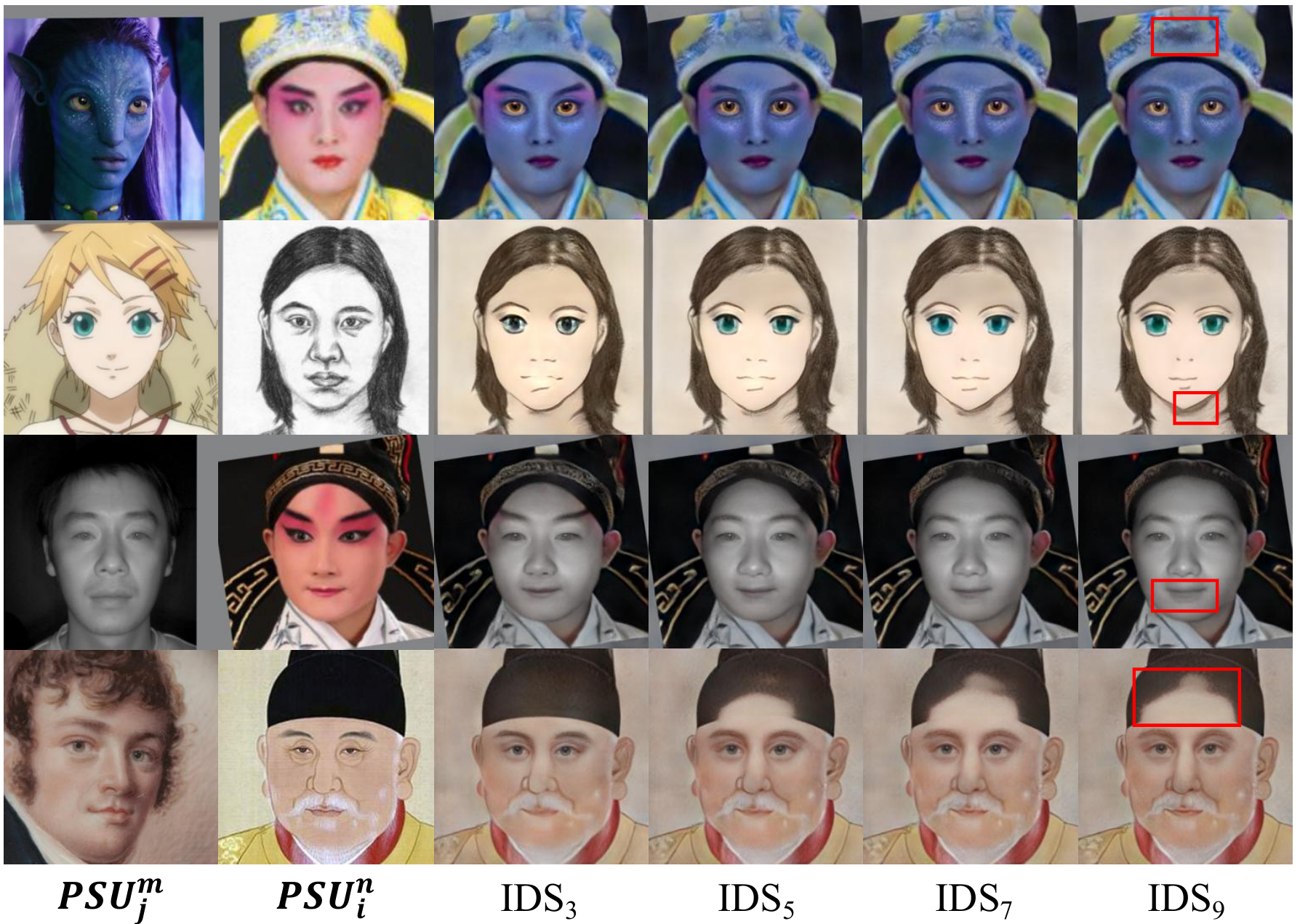}
\end{center}
\vspace{-13pt}
   \caption{Visual comparison of Styleverse on 30, 50, 70 and 90 thousand iterations, which demonstrates the phenomenon of PSU collapse.}
\label{fig:time}
\vspace{-11pt}
\end{figure}

\paragraph{PSU collapse} 
Mode collapse often occurs in GAN training, whereas Styleverse has PSU collapse when simultaneously handling the style transfer and content preservation. The quantitative metric curves of Styleverse are shown in supplementary material, as iteration goes by, the visual quality and quantitative values of some PSU transfer results become deteriorated. As for the visual comparison, there are some artifacts Of $IDS_{9}$ in raws 1\&4, Figure \ref{fig:time}. The gap of styled and original jaw in raw 2 becomes increasingly larger. The lip of the created NIR face based on Beijing Opera becomes blurred. As for $IDS_{3}$, the transferred style usually is inadequate. 

To make a reasonable quantitative comparison only for Styleverse, we select the metric values of $IDS_{7}$ presented in Table \ref{tab: nb}. The contents of Sketch, NIR and Beijing Opera are more easily transferred to other PSUs. Moreover, we calculate the average values of each column in Table \ref{tab: nb}, and find that the created Avatars and sketch faces based on other PSU contents are higher fidelity and quality. Note that there are supposed to be $M^{2}$ for $PSU_{ij}$, we calculate the quantitative values based on $PSU_{ij}^{tt}$ where $t$ means the same image index.
\begin{figure}[t]
\begin{center}
\includegraphics[width=0.7\linewidth]{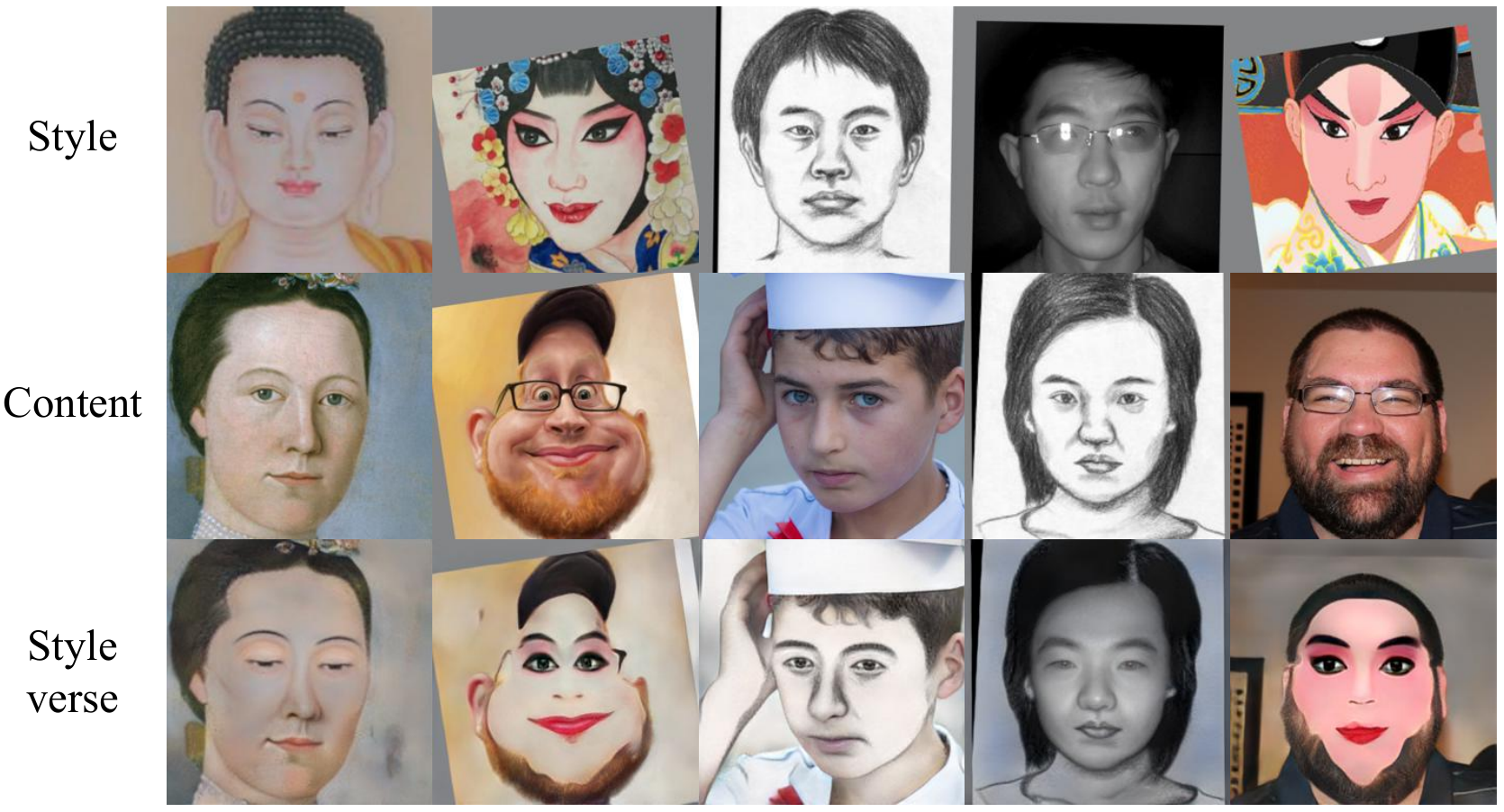}
\end{center}
\vspace{-13pt}
   \caption{Some hard cases of Styleverse.}
\label{fig:hard}
\vspace{-10pt}
\end{figure}

\subsection{Comparison with Other HIG Methods}
\begin{figure}[t]
\begin{center}
\includegraphics[width=0.7\linewidth]{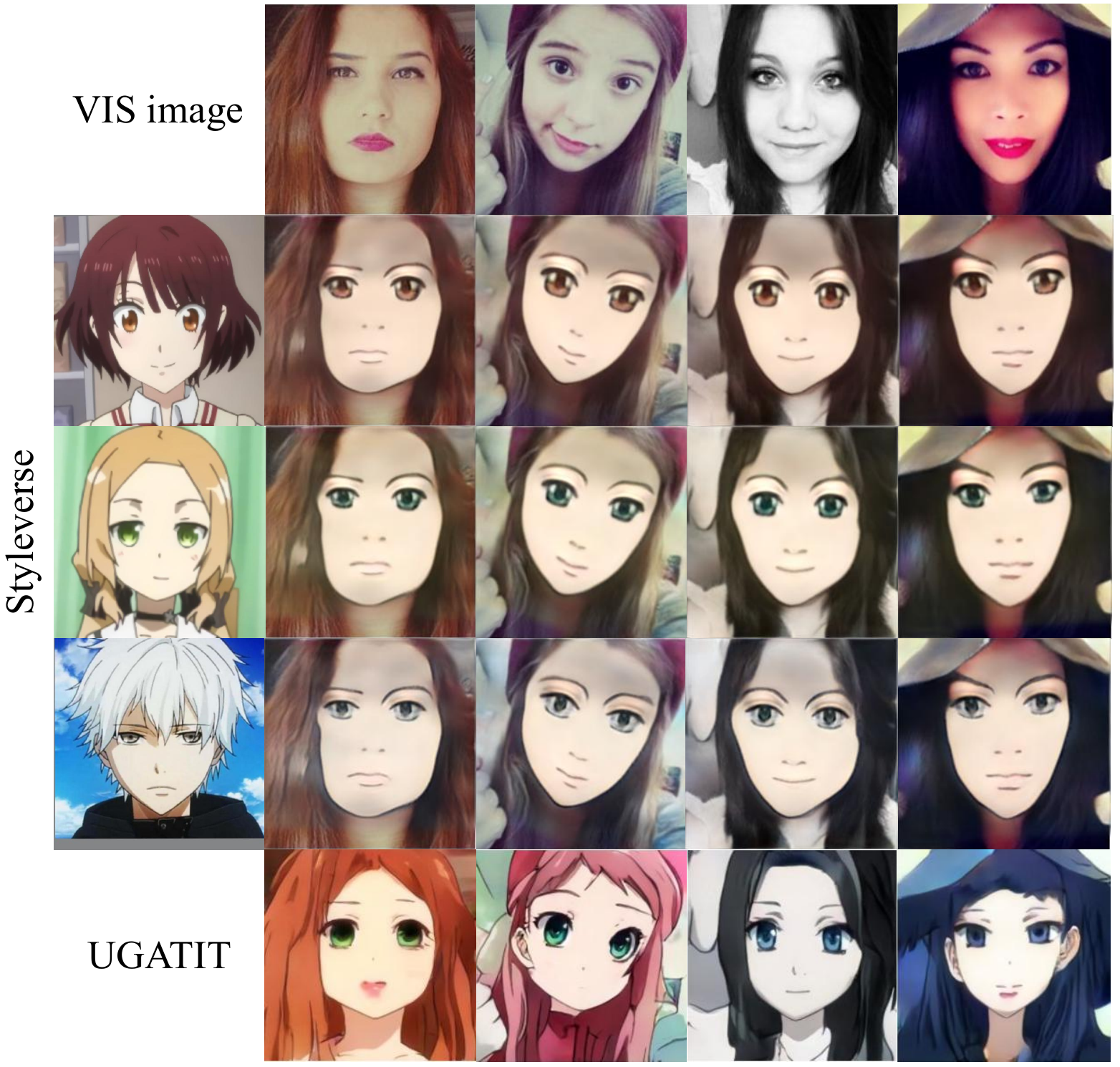}
\end{center}
\vspace{-13pt}
   \caption{Visual comparison of Styleverse and UGATIT \cite{ugatit}. Styleverse preserves the original face shape and background well, and achieves diverse synthesis of eyes. UGATIT only has one HIG result with smaller face areas. Styleverse relaxes the domain distribution of 2D Japanese anime that usually represents the character using narrowed jaw. Note that Styleverse only has one generator, while UGATIT has two generators, four discriminators and 6 auxiliary classifiers only for VIS-Anime HIG.}
\label{fig:uga}
\vspace{-10pt}
\end{figure}

As shown in Figure \ref{fig:uga}, we compare Styleverse with other state-of-the-art FST methods UGATIT \cite{ugatit} (More comparisons in supplementary material). UGATIT employs 2 generators and 4 discriminators to learn the dual mapping between the real-world and anime domains.  Our Styleverse is capable of generating diverse IDS faces, as shown in Figure \ref{fig:matrix}. Moreover, the created anime and VIS faces are competitive and controllable in Figure \ref{fig:uga}, even with only one generator and no discriminator.
\subsection{Limitations}
Styleverse first addresses the IDS task, and the Styleverse matrix contains lots of possibilities of new created faces. There are still some hard cases, as shown in Figure \ref{fig:hard}. Some aspects need to be further improved, \eg, the background distortion in heterogeneous domains (cols 1\&4), the facial occlusions (cols 2\&5), and  insufficient style transfer in local areas (col 3). 
\subsection{Broader impact}
Styleverse first studies Metaverse of identities in heterogeneous domains. Actually, the proposed highly topology-aware Styleverse is a high-level concept. 3D Styleverse, other image-level Styleverse of human or natural objects are also deserved to be studied.
\section{Conclusion}
We have proposed Styleverse to implement the topology-aware identity stylization based on the new established FS13 dataset, achieving high fidelity content and diverse styles. We first study and analyse this challenging IDS task using the effective heterogeneous-network-based Styleverse from an universal and lightweight perspective. In addition, we establish the first quantitative benchmark and the visual Styleverse matrix for the further research. Extensive comparison and ablation experiments demonstrate that our approach can handle high-quality and domain-aware IDS across heterogeneous domains.

{\small
\bibliographystyle{ieee_fullname}
\bibliography{egbib}
}

\clearpage
\section*{Supplementary material}
\paragraph{Contribution} \textbf{\emph{Lightweight and universal intelligence is the future research direction of highly topology-aware IDS task.}} Styleverse first integrates a large amount of facial heterogeneous domains and studies their mutual transformation. We conduct the highly topology-aware identity stylization considering controllable style transfer of the content. Styleverse first uses an universal generator to deal with the challenging and meaningful IDS problem based on the proposed dataset FS13.  
\begin{figure}[h]
\begin{center}
\includegraphics[width=1\linewidth]{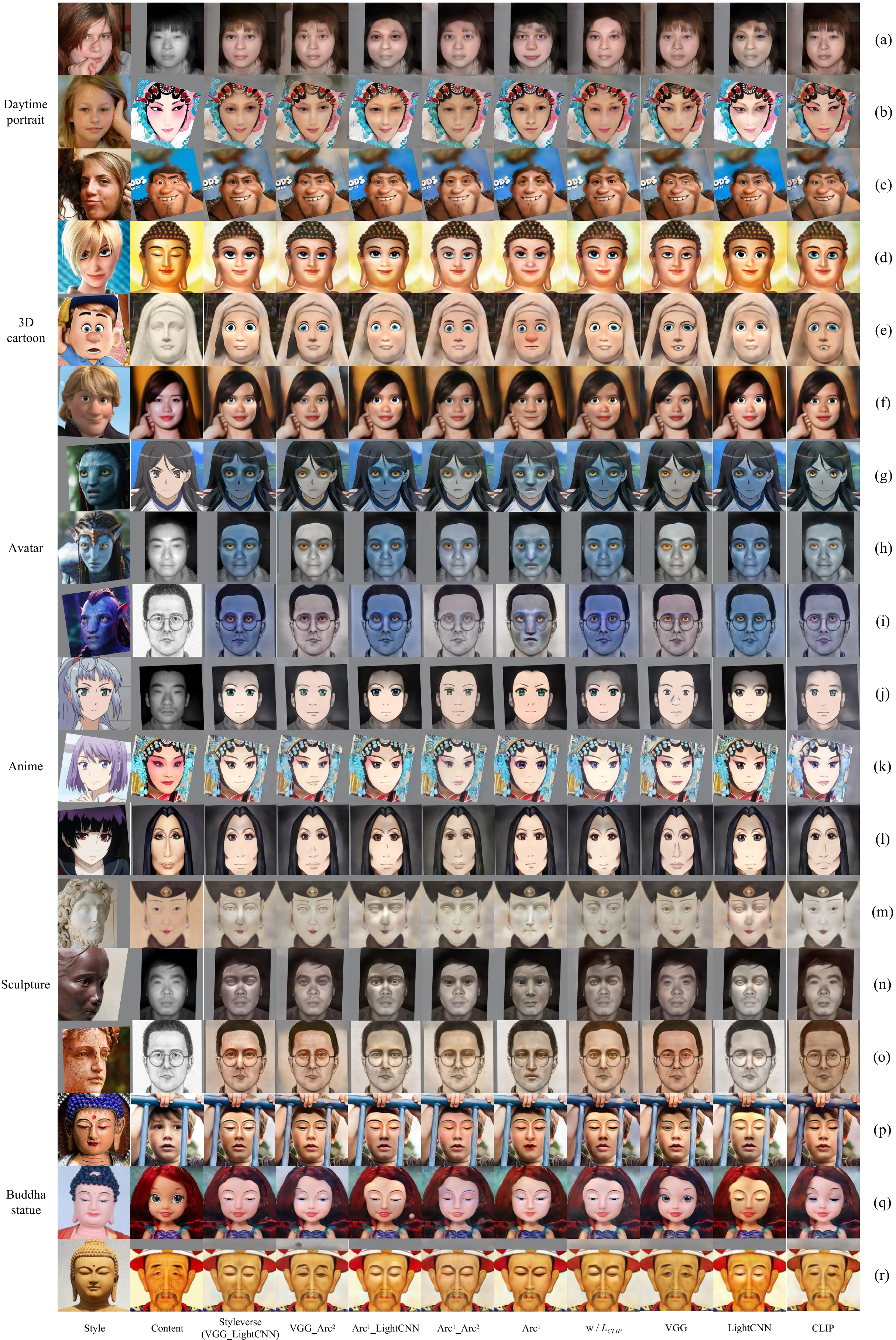}
\end{center}
\vspace{-13pt}
   \caption{Visual IDS comparison of Styleverse and the variants based on Arcface \cite{arcface2019}, VGG and CLIP. There are examples for daytime portrait, 3D cartoon, Avatar, anime, sculpture and Buddha statue.}
\label{fig:arc}
\vspace{-10pt}
\end{figure}
\begin{figure}[h]
\begin{center}
\includegraphics[width=0.99\linewidth]{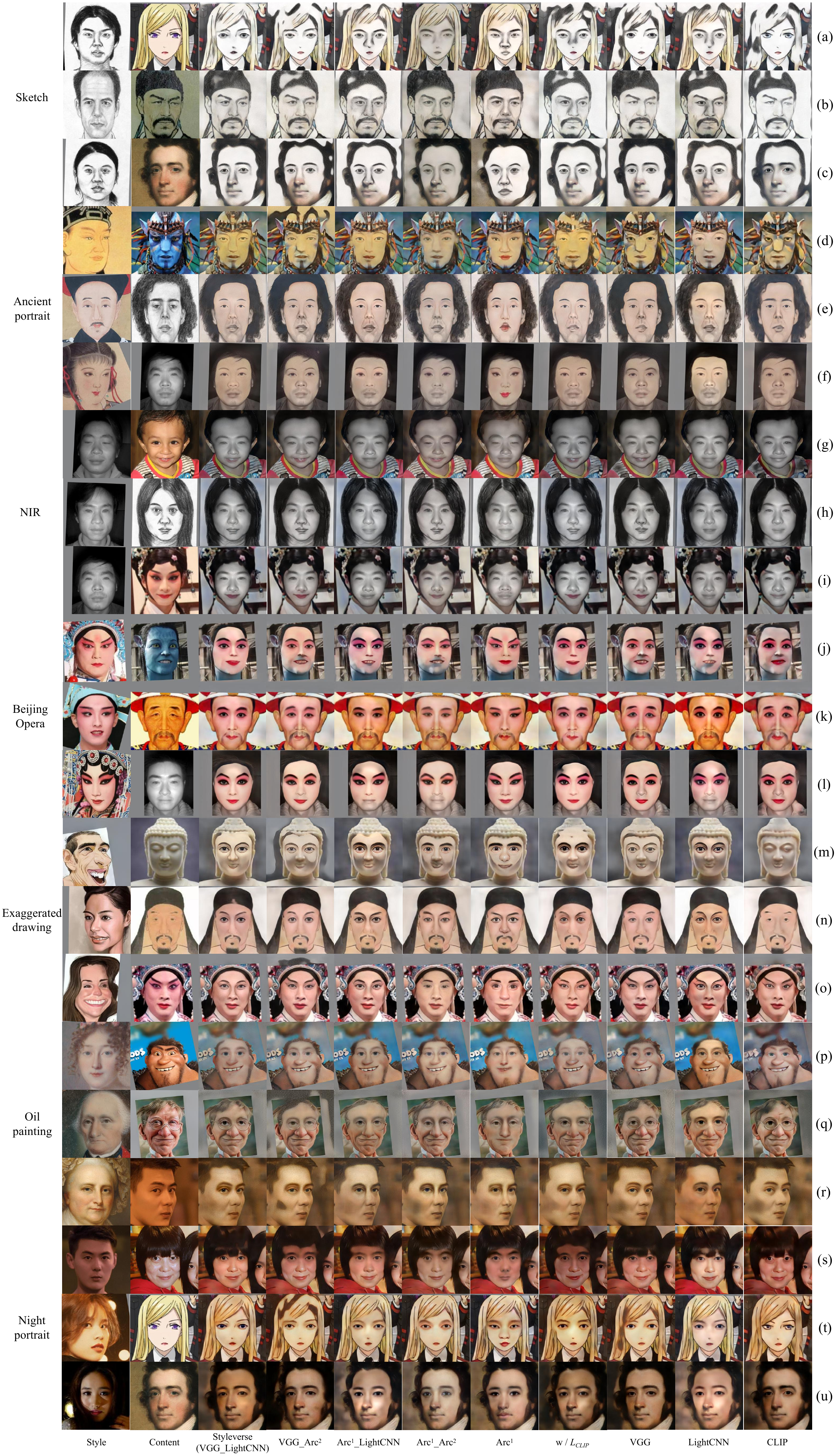}
\end{center}
\vspace{-13pt}
   \caption{Visual IDS comparison of Styleverse and the variants based on Arcface \cite{arcface2019}, VGG and CLIP. There are examples for sketch, ancient portrait, NIR, Beijing Opera, exaggerated drawing, oil painting and night portrait.}
\label{fig:arc3}
\vspace{-10pt}
\end{figure}
   

\begin{table*}[h]
\caption{Quantitative evaluation of ablation study with FID, KID$\times100$,NIQE, $ID_{style}$ and $ID_{content}$. The red values mean the top-1 scores for each row, among Styleverse (VGG\_LightCNN), only VGG, only LightCNN, and only CLIP.}
\vspace{-10.pt}
\scalebox{0.5}{
\begin{tabular}{|c|cccc|cccc|cccc|cccc|cccc|}
\hline 
Dataset & \multicolumn{4}{c|}{FID$\downarrow$} & \multicolumn{4}{c|}{KID$\downarrow$} & \multicolumn{4}{c|}{NIQE$\downarrow$} & \multicolumn{4}{c|}{$ID_{style}\downarrow$}& \multicolumn{4}{c|}{$ID_{content}\downarrow$}\tabularnewline
\hline 
Methods & Styleverse & VGG & LightCNN & CLIP & Styleverse & VGG & LightCNN & CLIP &Styleverse & VGG & LightCNN & CLIP &Styleverse & VGG & LightCNN & CLIP &Styleverse & VGG & LightCNN & CLIP\tabularnewline
 
\hline 
Daytime portrait & 181.84&	187.28&	\textbf{\textcolor{red}{178.86}}&	189.5&	6.8&	6.69&	\textbf{\textcolor{red}{6.59}}&	7.65&	7.75&	\textbf{\textcolor{red}{7.66}}&	7.79&	8.4&	\textbf{\textcolor{red}{0.86}}&	0.88&	0.88&	0.89&	0.44&	0.23&	0.38&	\textbf{\textcolor{red}{0.17}}
\tabularnewline
3D cartoon & \textbf{\textcolor{red}{146.56}}&149.49&	149.49&	153.5&	3.46&	\textbf{\textcolor{red}{3.43}}&	3.97&	4.64&	\textbf{\textcolor{red}{7.65}}&	7.7&	7.73&	8.17&	\textbf{\textcolor{red}{0.67}}&	0.75&	0.72&	0.73&	0.48&	0.29&	0.41&	\textbf{\textcolor{red}{0.28}}
\tabularnewline
Avatar & \textbf{\textcolor{red}{111.54}}&123.87&	114.91&	122.02&	\textbf{\textcolor{red}{3.95}}&	4.69&	3.98&	4.42&	\textbf{\textcolor{red}{7.72}}&	7.73&	7.77&	8&	\textbf{\textcolor{red}{0.75}}&	0.81&	0.79&	0.8&	0.43&	0.24&	0.35&	\textbf{\textcolor{red}{0.24}}
\tabularnewline
Anime & \textbf{\textcolor{red}{147.19}}&	148.17&	151.87&	149.97&	6.35&	\textbf{\textcolor{red}{6.11}}&	6.74&	6.36&	\textbf{\textcolor{red}{7.63}}&	7.68&	7.63&	7.95&	\textbf{\textcolor{red}{0.62}}&	0.73&	0.63&	0.7&	0.55&	\textbf{\textcolor{red}{0.36}}&	0.5&	0.4
\tabularnewline
Sculpture & \textbf{\textcolor{red}{144.34}}&	154.44&	149.31&	156.35&	\textbf{\textcolor{red}{5.43}}&	6.1&	5.64&	6.18&	\textbf{\textcolor{red}{7.98}}&	8.05&	8.01&	8.52&	\textbf{\textcolor{red}{0.77}}&	0.83&	0.81&	0.84&	0.48&	0.25&	0.38&	\textbf{\textcolor{red}{0.16}}
\tabularnewline
Buddha statue & \textbf{\textcolor{red}{138}}&147.14&	142.41&	148.78&	\textbf{\textcolor{red}{6.6}}&	6.96&	7.17&	7.91&	\textbf{\textcolor{red}{7.88}}&	7.95&	7.97&	8.44&	\textbf{\textcolor{red}{0.6}}&	0.71&	0.65&	0.69&	0.50& 	0.29&	0.44&	\textbf{\textcolor{red}{0.29}}
\tabularnewline
Sketch & 203.24&\textbf{\textcolor{red}{201.83}}&	202.04&	204.54&	20.42&	\textbf{\textcolor{red}{19.38}}&	20.29&	21.05&	\textbf{\textcolor{red}{7.35}}&	7.37&	7.55&	7.72&	\textbf{\textcolor{red}{0.68}}&	0.75&	0.70& 	0.81&	0.52&	0.35&	0.48&	\textbf{\textcolor{red}{0.24}}
\tabularnewline
Ancient portrait & 152.5&\textbf{\textcolor{red}{145.21}}&	156.48&	152.96&	5.01&	\textbf{\textcolor{red}{3.81}}&	4.83&	4.64&	\textbf{\textcolor{red}{7.98}}&	8.05&	8.07&	8.58&	\textbf{\textcolor{red}{0.65}}&	0.72&	0.71&	0.74&	0.51&	0.33&	0.44&	\textbf{\textcolor{red}{0.23}}
\tabularnewline
NIR & 207.47&\textbf{\textcolor{red}{198.14}}&	205.37&	202.63&	17.2&	\textbf{\textcolor{red}{16.61}}&	17.34&	17.49&	\textbf{\textcolor{red}{7.69}}&	7.71&	7.76&	7.94&	\textbf{\textcolor{red}{0.68}}&	0.77&	0.69&	0.72&	0.62&	\textbf{\textcolor{red}{0.43}}&	0.65&	0.53
\tabularnewline
Beijing Opera & 125.39&\textbf{\textcolor{red}{124.57}}&	128.07&	125.37&	5.18&	\textbf{\textcolor{red}{4.41}}&	4.97&	5.22&	\textbf{\textcolor{red}{7.76}}&	7.81&	7.83&	8.27&	\textbf{\textcolor{red}{0.67}}&	0.74&	0.72&	0.72&	0.54&	\textbf{\textcolor{red}{0.34}}&	0.48&	0.37
\tabularnewline
Exaggerated drawing &129.58&\textbf{\textcolor{red}{128.86}}&	131.81&	139.11&	4.17&	\textbf{\textcolor{red}{3.89}}&	4.39&	4.77&	\textbf{\textcolor{red}{7.52}}&	7.55&	7.64&	8.18&	\textbf{\textcolor{red}{0.83}}&	0.83&	0.84&	0.85&	0.43&	0.24&	0.37&	\textbf{\textcolor{red}{0.16}}
\tabularnewline
Oil painting & 157.89&158.58&	\textbf{\textcolor{red}{155.18}}&	162.59&	7.36&	\textbf{\textcolor{red}{7.03}}&	7.15&	7.92&	7.85&	\textbf{\textcolor{red}{7.83}}&	7.99&	8.34&	\textbf{\textcolor{red}{0.78}}&	0.83&	0.80 &	0.83&	0.49&	0.30 &	0.45&	\textbf{\textcolor{red}{0.25}}
\tabularnewline
Night portrait & 148.83&149.69&	\textbf{\textcolor{red}{147.31}}&	154.77&	4.28&	4.29&	\textbf{\textcolor{red}{3.74}}&	5.17&	\textbf{\textcolor{red}{7.93}}&	7.93&	8&	8.42&	\textbf{\textcolor{red}{0.83}}&	0.86&	0.84&	0.87&	0.48&	0.26&	0.39&	\textbf{\textcolor{red}{0.16}}
\tabularnewline
\hline 
Mean & \textbf{\textcolor{red}{153.41}}&155.17&	154.85&	158.62&	\textbf{7.40}& 	\textbf{\textcolor{red}{7.18}}&	7.44&	7.95&	\textbf{\textcolor{red}{7.74}}&	7.77&	7.82&	8.22&	\textbf{\textcolor{red}{0.72}}&	0.78&	0.75&	0.78&	\textbf{0.49}&	0.30 &	0.44&	\textbf{\textcolor{red}{0.26}}
\tabularnewline
\hline
\end{tabular}
}
\vspace{-10.pt}
\label{tab: ablation_supp}
\end{table*}
\begin{table*}[h]
\caption{Quantitative evaluation of ablation study with mean scores of FID, KID$\times100$,NIQE, $ID_{style}$ and $ID_{content}$. The red values mean the top-1 scores, among Styleverse, VGG\_$Arc^{2}$, $Arc^{1}\_LightCNN$, $Arc^{1}\_Arc^{2}$, $Arc^{1}$, w/ $\mathcal{L}_{CLIP}$, only VGG, LightCNN and CLIP.}
\vspace{-10.pt}
\scalebox{0.225}{
\begin{tabular}{|c|ccccccccc|ccccccccc|ccccccccc|ccccccccc|ccccccccc|}
\hline 
Dataset & \multicolumn{9}{c|}{FID$\downarrow$} & \multicolumn{9}{c|}{KID$\downarrow$} & \multicolumn{9}{c|}{NIQE$\downarrow$} & \multicolumn{9}{c|}{$ID_{style}\downarrow$}& \multicolumn{9}{c|}{$ID_{content}\downarrow$}\tabularnewline
\hline 
\makecell[c]{Methods} & \makecell[c]{Styleverse} & \makecell[c]{VGG\_$Arc^{2}$} & \makecell[c]{$Arc^{1}$\_\\LightCNN} &\makecell[c]{$Arc^{1}$\_$Arc^{2}$} & \makecell[c]{$Arc^{1}$} & \makecell[c]{w/ $\mathcal{L}_{CLIP}$} & VGG & LightCNN & CLIP&  \makecell[c]{Styleverse} & \makecell[c]{VGG\_$Arc^{2}$} & \makecell[c]{$Arc^{1}$\_\\LightCNN} &\makecell[c]{$Arc^{1}$\_$Arc^{2}$} & \makecell[c]{$Arc^{1}$} & \makecell[c]{w/ $\mathcal{L}_{CLIP}$} & VGG & LightCNN & CLIP&\makecell[c]{Styleverse} & \makecell[c]{VGG\_$Arc^{2}$} & \makecell[c]{$Arc^{1}$\_\\LightCNN} &\makecell[c]{$Arc^{1}$\_$Arc^{2}$} & \makecell[c]{$Arc^{1}$} & \makecell[c]{w/ $\mathcal{L}_{CLIP}$} & VGG & LightCNN & CLIP&\makecell[c]{Styleverse} & \makecell[c]{VGG\_$Arc^{2}$} & \makecell[c]{$Arc^{1}$\_\\LightCNN} &\makecell[c]{$Arc^{1}$\_$Arc^{2}$} & \makecell[c]{$Arc^{1}$} & \makecell[c]{w/ $\mathcal{L}_{CLIP}$} & VGG & LightCNN & CLIP&\makecell[c]{Styleverse} & \makecell[c]{VGG\_$Arc^{2}$} & \makecell[c]{$Arc^{1}$\_\\LightCNN} &\makecell[c]{$Arc^{1}$\_$Arc^{2}$} & \makecell[c]{$Arc^{1}$} & \makecell[c]{w/ $\mathcal{L}_{CLIP}$} & VGG & LightCNN & CLIP\tabularnewline
 
\hline 
Mean & \textbf{153.41}&	155.68&	151.09&	154.25&	\textbf{\textcolor{red}{148.49}}&	155.25&	155.17&	154.85&	158.62&
\textbf{7.40} &	7.36&	7.15&	7.06&	\textbf{\textcolor{red}{6.65}}&	7.75& 7.18&	7.44&	7.95&	\textbf{\textcolor{red}{7.74}}&	7.87&	7.77&	7.82&	7.89&	7.90 &7.77&	7.82&	8.22&	\textbf{0.72}&	0.76&	0.68&	0.70& 	\textbf{\textcolor{red}{0.4}}&	0.73&0.78&	0.75&	0.78&
	\textbf{0.49}&	0.37&	0.51&	0.42&	0.63&	0.47
&0.30& 	0.44&	\textbf{\textcolor{red}{0.26}}
\tabularnewline
\hline
\end{tabular}
}
\vspace{-16.pt}
\label{tab: ablation}
\end{table*}

\begin{table}[h]
\centering
\caption{IDS performance analysis of Styleverse.}
\scalebox{0.4}{
\begin{tabular}{|c|ccc|ccc|}
\hline 
\makecell[c]{PSUs} & \makecell[c]{top-1} & \makecell[c]{top-2} & top-3 & FID & KID & NIQE\tabularnewline
\hline 
Daytime portrait & 2&	2&	5 &207.08 &11.39 &7.71\tabularnewline
3D cartoon & 2&	1&	3 &159.70 &6.67 &7.66\tabularnewline
Avatar & 2&	0&	2& \textbf{122.52} & \textbf{6.13} & 7.71\tabularnewline
Anime & 2&	0&	3 &162.31 &9.74 &7.59\tabularnewline
Sculpture & 2&	4&	2&167.42 &9.70 &7.97\tabularnewline
Buddha statue & 2&	3&	1& 152.09 &9.92 &7.87\tabularnewline
Sketch & \textbf{12}&	4&	2 &215.57 &24.27 &\textbf{7.30}\tabularnewline
Ancient portraits & 2&	3&	3&170.09 &8.36 &7.98\tabularnewline
NIR & 4&	\textbf{10}&	1 &228.04 &23.12 &7.71\tabularnewline
Beijing Opera & 2&	3&	4 &137.60 &7.97 &7.76\tabularnewline
Exaggerated drawing & 3&	4&	\textbf{7} &151.38 &8.09 &7.52\tabularnewline
Oil painting & 2&	5&	3&185.35 &12.52 &7.84\tabularnewline
Night portrait & 2&	0&	3 &166.16 &7.94 &7.91\tabularnewline
\hline
\end{tabular}
}
\vspace{-8.pt}
\label{tab:two}
\end{table}
\begin{figure}[h]
\begin{center}
\includegraphics[width=0.5\linewidth]{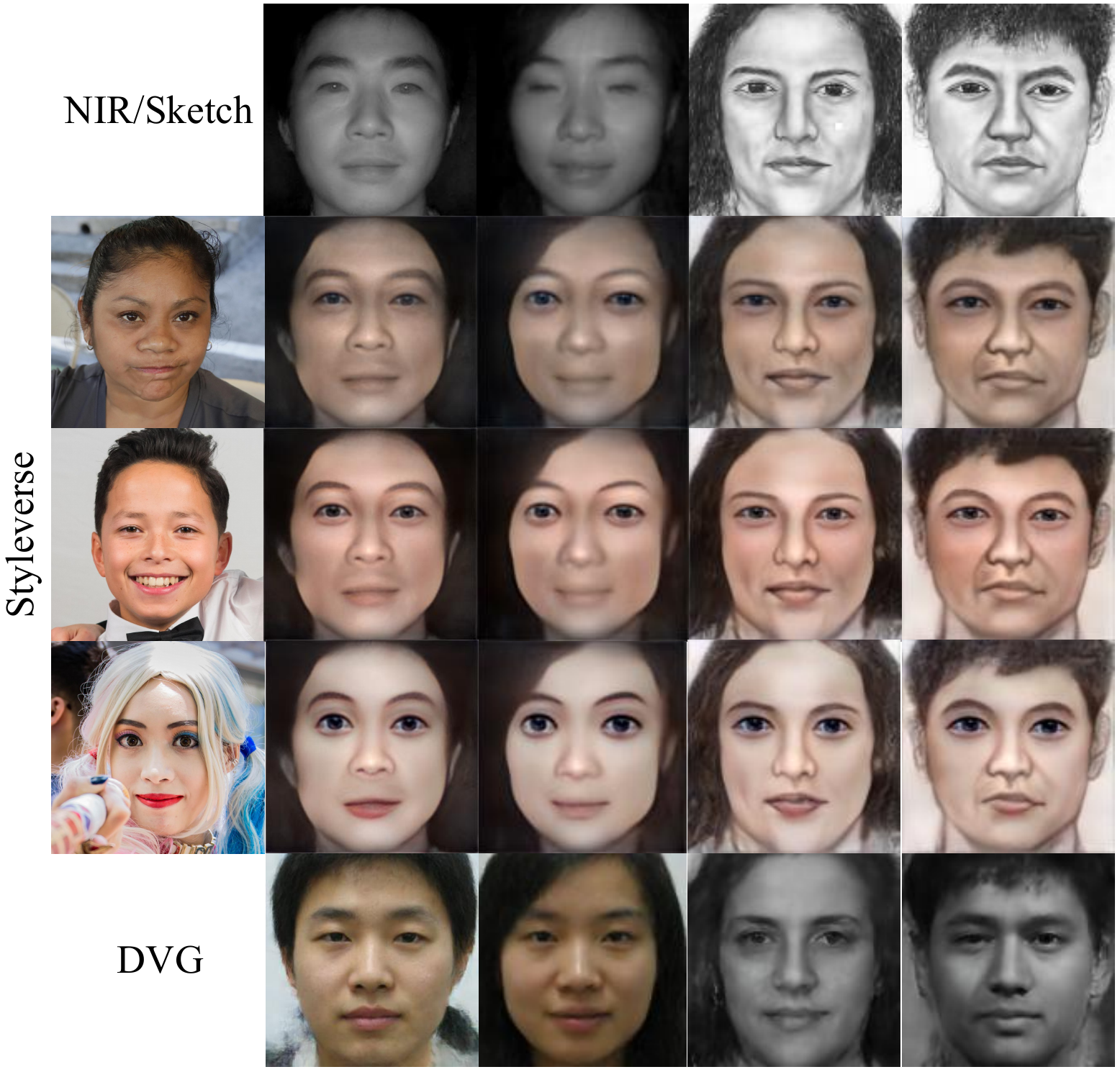}
\end{center}
\vspace{-13pt}
   \caption{Visual IDS comparison of Styleverse and DVG \cite{fu2019dual}. Styleverse achieves diverse NIR-VIS or Sketch-VIS heterogeneous image generation with lighting style preservation. Note that our styleverse uses a small subset of FFHQ \cite{stylegan19} rather than CASIA NIR-VIS \cite{casia_nir_vis2} in DVG for the VIS domain.}
\label{fig:nir}
\vspace{-10pt}
\end{figure}
\begin{figure*}[h]
\begin{center}
\includegraphics[width=1\linewidth]{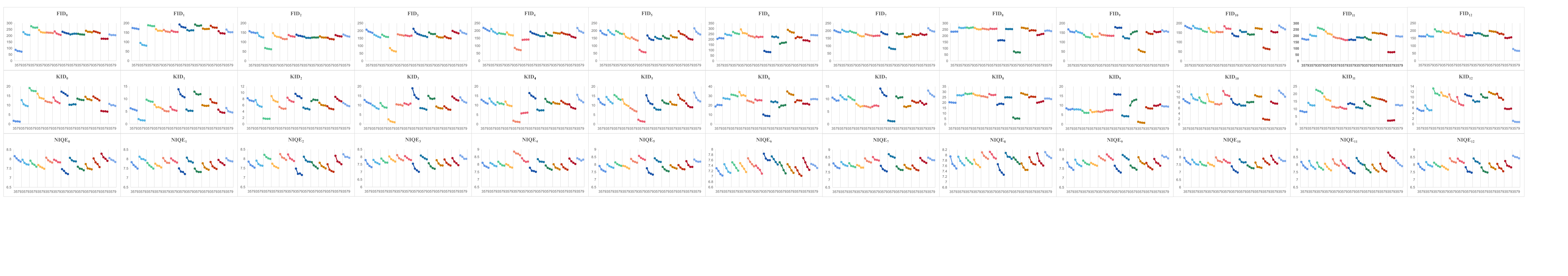}
\end{center}
\vspace{-13pt}
   \caption{The quantitative metric curves of Styleverse. For each unit, e.g., $FID_{0}$, we record corresponding evaluation metrics on 30, 50, 70 and 90 thousand iterations, where different color represents different PSU content. From left to right, these segmented curves indicate daytime portrait, 3D cartoon, Avatar, anime, sculpture, Buddha statue, sketch, ancient portrait, NIR, Beijing Opera, exaggerated drawing, oil painting and night portrait. Moreover, for each row, 0-12 also means this PSU order.}
\label{fig:total}
\vspace{-10pt}
\end{figure*}

\begin{figure*}[h]
\begin{center}
\includegraphics[width=1\linewidth]{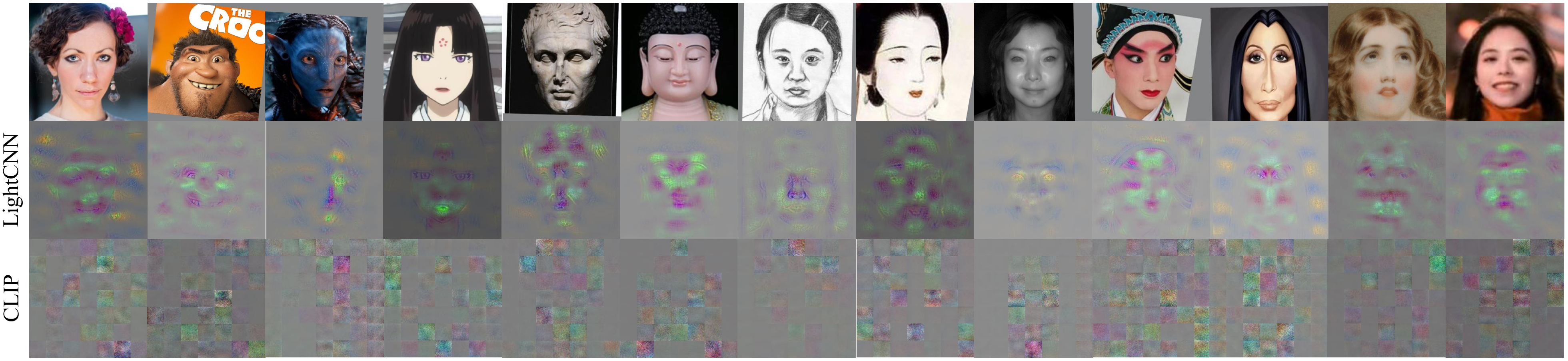}
\end{center}
\vspace{-13pt}
   \caption{Guided backpropagation (GBP) \cite{gbp} of LightCNN and CLIP that uses VisualTransformer as the image decoder with $patch\_size = 32$ and $vision\_width = 768$. GBP maps of LightCNN focus more on the topological face area.}
\label{fig:clip_cam}
\vspace{-10pt}
\end{figure*}
For the first time, we have systematically explored and analyzed IDS from an universal and lightweight perspective. IDS is different from the previous style transfer task that mainly transfers the textural feature \cite{WCT,aams,avatar,adain2017}. Moreover, it is more universal and lightweight than other GAN-based heterogeneous face generation methods \cite{ugatit,cyclegan} or VAE-based methods \cite{fu2019dual}.

We propose an effective heterogeneous-network-based IDS framework. We analyse different Styleverse variants, \eg, w/ Arcface, w/ $\mathcal{L}_{CLIP}$, and compare with several sota style transfer methods. We carefully establish the first IDS quantitative benchmark (Table 1, 2, 3 in main manuscript and Table \ref{tab: ablation_supp}). 

Extensive comparison and ablation experiments demonstrate that our approach can handle high-quality and domain-aware IDS across heterogeneous domains. This work is helpful for the further study of IDS. We provide more studies of ablation study and analysis of Styleverse as follows.

Note that if achieving $N$-domain IDS, other HIG methods, \eg, UGATIT, need to train  $C_{N}^{2}$ times, \eg, 78 times for UGATIT when $N$=13, which will rely on a large amount of computation. It is unreasonable and unfair to compare all models based on UGATIT for cross-domain mutual IDS. While our Styleverse handles all cross-domain IDS subtasks using a single generator, as shown in our Styleverse matrix (Figure 2 in the main manuscript).
\paragraph{Additional ablation study}
We conduct the ablation study as shown in Figure \ref{fig:arc} and Figure \ref{fig:arc3}.  Note that all Styleverse variants use $\mathcal{L}_{rec}$, $\mathcal{L}_{CCX}$ and $\mathcal{L}_{SCX}$ with the same weights of Styleverse, but different $\mathcal{L}_{PSU}$ losses.
$Arc^{1}$ means the original instance-level identity recognition network, and $Arc^{2}$ is the pretrained PSU recognition network based on FS13.

\noindent\bm{$VGG\_Arc^{2}$} replaces the LightCNN of Styleverse, which has worse $ID_{style}$ score. It indicates that LightCNN is better at extracting distinctive PSU styles than Arcface. It's results are not vivid enough, \eg, (g, h, i) for Avatar IDS in Figure \ref{fig:arc}.

\noindent\bm{$Arc^{1}\_LightCNN$} replaces the VGG of Styleverse using Arcface whose ability of capturing textural feature of style image is limited, \eg, (n, u) in Figure \ref{fig:arc3}.

\noindent\bm{$Arc^{1}\_Arc^{2}$} has some artifacts (\eg, hair in (a), Figure \ref{fig:arc}) because of constraint of Arcface, and conducts insufficient IDS, \eg, the mouth is not the anime style in (k, l) of Figure \ref{fig:arc}.

\noindent\bm{$Arc^{1}$} only uses the identity embedding from Arcface to modulate Styleverse, which results in transferring more identity of the style face, \eg, (e-i) in Figure \ref{fig:arc}, (d, f) in Figure \ref{fig:arc3}. This makes IDS ignore the original expression of the content, \eg, (l) in Figure \ref{fig:arc}, (g, i, s) in Figure \ref{fig:arc3}. We train this variant using identity consistency loss as follows
\begin{equation}
\mathcal{L}_{PSU}=1 -\langle z_{Arcface}(PSU_{ij}^{nm}) ,z_{Arcface}\left(PSU_{j}^{m}\right) \rangle,
\end{equation}
where $\langle\cdot, \cdot\rangle$ means cosine similarity.

\noindent\bm{$w/ \mathcal{L}_{CLIP}$} combines a Contrastive Language-Image Pretraining (CLIP) \cite{clip} perceptual loss to constrain Styleverse in response to the heterogeneous-domain text prompt. We first finetune the pretrained CLIP model using FS13 and the corresponding texts, \eg, 'Beijing Opera', '3D cartoon'. $\mathcal{L}_{CLIP}$ is formulated as
\begin{equation}
\mathcal{L}_{CLIP}=D_{CLIP}(PSU_{ij}^{nm},t),
\end{equation}
where $D_{CLIP}$ means the cosine distance between the CLIP embeddings of its image and text features. This variant improves Styleverse in some cases, \eg, (d, g, q) in Figure \ref{fig:arc}, where the boundary areas of face and hair are more natural. 

\noindent\textbf{Only VGG} uses the feature embedding from adaptive average pooling of VGG $conv4\_1$ to modulate Styleverse. The loss is formulated as
\begin{equation}
\mathcal{L}_{PSU}=1 -\langle z_{VGG}(PSU_{ij}^{nm}) ,z_{VGG}\left(PSU_{j}^{m}\right) \rangle,
\end{equation}
where $\langle\cdot, \cdot\rangle$ means cosine similarity. This variant imposes weak heterogeneous-aware style transfer to the content, \eg, (e, j, l, q, r) in Figure \ref{fig:arc}.

\noindent\textbf{Only LightCNN} uses the heterogeneous-aware embedding from LightCNN to modulate Styleverse. The textural information is not transferred as shown in (o, q, r) of Figure \ref{fig:arc}, (k, p, s) of Figure \ref{fig:arc3}.

\noindent\textbf{Only CLIP} modulates Styleverse only using the feature from image encoder of the pretrained CLIP. It has potential for achieving IDS in some heterogeneous domains such as Anime in Figure \ref{fig:arc}, ancient portrait in Figure \ref{fig:arc3}. This variant also uses $\mathcal{L}_{CLIP}$.

\noindent\textbf{Styleverse} conducts more visually controllable IDS by considering textural and distinctive PSU style transfer, as for the qualitative comparison in Figure \ref{fig:arc} and \ref{fig:arc3}. 

\paragraph{Guided backpropagation}
As shown in Figure \ref{fig:clip_cam},  we compare the GuidedBackPropagation maps of LightCNN and CLIP model where there are $7\times7$ patches. GBP maps of LightCNN focus more on the topological face area. Other GBP maps are shown in Figure 11 of the main manuscript.
\paragraph{Analysis of PSU collapse}
Furthermore, we analyse the statistical quantities of Table 3 in the main manuscript, as shown in Table \ref{tab:two}. Specifically, we count the number of top-1, top-2 and top-3 scores in the horizontal direction of Table 3 where the contents of Sketch, NIR and Beijing Opera are more easily transferred to other PSUs. Moreover, we calculate the average values of each column in Table 3, and find that the created Avatars and sketch faces based on other PSU contents are higher fidelity and quality. The quantitative metric curves of Styleverse are shown in Figure \ref{fig:total}. PSU collapse happens in around 70 thousand iteration. The challenging and meaningful IDS task deserves to be further studied, which is helpful to exploit the Metaverse of identities in diverse parallel style universes.

\end{document}